\documentclass[10pt,twocolumn,letterpaper]{article}

\usepackage[pagenumbers]{cvpr} 

\usepackage{graphicx}
\usepackage{amsmath}
\usepackage{amssymb}
\usepackage{booktabs}

\usepackage{times}
\usepackage{epsfig}
\usepackage{multirow}

\usepackage{enumitem}
\usepackage{bbm}
\usepackage{amsthm}
\usepackage{dsfont}
\usepackage{bm}

\usepackage{mathtools}

\usepackage{algpseudocode}
\usepackage{algorithm}

\usepackage{caption}
\usepackage{subcaption}
\usepackage{xcolor}
\usepackage[pagebackref,breaklinks,colorlinks,bookmarks=false,urlcolor=black,citecolor={green!60!black}]{hyperref}

\makeatletter
\def\ps@myheadings{%
    \let\@oddfoot\@empty\let\@evenfoot\@empty
    \def\@evenhead{\thepage\hfil\slshape\leftmark}%
    \def\@oddhead{{\slshape\rightmark}\hfil\thepage}%
    \let\@mkboth\@gobbletwo
    \let\sectionmark\@gobble
    \let\subsectionmark\@gobble
    }
  \if@titlepage
  \renewcommand\maketitle{\begin{titlepage}%
  \let\footnotesize\small
  \let\footnoterule\relax
  \let \footnote \thanks
  \null\vfil
  \vskip 60\p@
  \begin{center}%
    {\LARGE \@title \par}%
    \vskip 3em%
    {\large
     \lineskip .75em%
      \begin{tabular}[t]{c}%
        \@author
      \end{tabular}\par}%
      \vskip 1.5em%
    {\large \@date \par}
  \end{center}\par
  \@thanks
  \vfil\null
  \end{titlepage}%
  \setcounter{footnote}{0}%
}
\else
\renewcommand\maketitle{\par
  \begingroup
    \renewcommand\thefootnote{\@fnsymbol\c@footnote}%
    \def\@makefnmark{\rlap{\@textsuperscript{\normalfont\color{black}\@thefnmark}}}%
    \long\def\@makefntext##1{\parindent 1em\noindent
            \hb@xt@1.8em{%
                \hss\@textsuperscript{\normalfont\@thefnmark}}##1}%
    \if@twocolumn
      \ifnum \col@number=\@ne
        \@maketitle
      \else
        \twocolumn[\@maketitle]%
      \fi
    \else
      \newpage
      \global\@topnum\z@   
      \@maketitle
    \fi
    \thispagestyle{plain}\@thanks
  \endgroup
  \setcounter{footnote}{0}%
}
\makeatother

\makeatletter
\newcommand\fs@nobottomruled{\def\@fs@cfont{\bfseries}\let\@fs@capt\floatc@ruled
  \def\@fs@pre{}
  \def\@fs@post{}
  \def\@fs@mid{\kern2pt\hrule\kern2pt}%
  \let\@fs@iftopcapt\iftrue}
\makeatother

\renewcommand{\dblfloatpagefraction}{0.99}

\renewcommand{\topfraction}{0.99} 
\renewcommand{\bottomfraction}{0.99} 
\renewcommand{\textfraction}{0.01} 
\renewcommand{\floatpagefraction}{0.99}

\usepackage{multibib}
\newcites{latex}{References}

\usepackage[capitalize]{cleveref}
\crefname{section}{Sec.}{Secs.}
\Crefname{section}{Section}{Sections}
\Crefname{table}{Table}{Tables}
\crefname{table}{Tab.}{Tabs.}

\DeclareSymbolFont{extraup}{U}{zavm}{m}{n}
\DeclareMathSymbol{\varheart}{\mathalpha}{extraup}{86}
\DeclareMathSymbol{\vardiamond}{\mathalpha}{extraup}{87}

\def\cvprPaperID{5094} 
\def\confName{CVPR}
\def\confYear{2022}

\newcommand{\comment}[1]{ }

\renewcommand\vec[1]{\ensuremath\boldsymbol{#1}}
\renewcommand\cdots{...}

\newcommand{\mZ}{\mathbf{Z}}

\newcommand{\vy}{\mathbf{y}}
\newcommand{\valpha}{\bm{\alpha}}

\newcommand{\tD}{\vec{\mathcal{D}}}

\newcommand{\mX}{\mathbf{X}}
\newcommand{\mA}{\mathbf{A}}

\newcommand{\mbrp}[1]{\mathbb{R}_{+}^{#1}}
\newcommand{\mbr}[1]{\mathbb{R}^{#1}}

\newcommand{\tAnb}{\mathcal{A}}

\newcommand{\idx}[1]{\mathcal{I}_{#1}}

\newcommand{\vpsi}{\boldsymbol{\psi}}

\newcommand{\mPsi}{\vec{\Psi}}

\DeclareMathOperator*{\softmingg}{SoftMin_{\bar{\gamma}}}
\DeclareMathOperator*{\softming}{SoftMin_\gamma}
\DeclareMathOperator*{\topminb}{TopMin_\beta}
\DeclareMathOperator*{\topmaxbb}{TopMax_{NZ\beta}}

\DeclareMathOperator*{\detach}{Detach}

\def\eg{\emph{e.g.}}

\newcommand{\mD}{\boldsymbol{D}}
\newcommand{\vd}{\boldsymbol{d}}

\newcommand{\stkout}[1]{{\ifmmode\text{\sout{\ensuremath{#1}}}\else\sout{#1}\fi}}

\begin{document}

\title{3D Skeleton-based Few-shot Action Recognition with JEANIE is not so Na\"ive\vspace{-0.3cm}}

\author{%
  Lei Wang$^{\dagger,\S}$, \quad Jun Liu$^{\spadesuit}$, \quad Piotr Koniusz\thanks{The corresponding author.}$\;^{,\S,\dagger}$\\\vspace{0.3cm}
  \!\!\!\!\!\!$^{\dagger}$The Australian National University \!\!\quad \!\!
  $^{\spadesuit}$Singapore University of Technology and Design \!\!\quad\!\! $^\S$Data61/CSIRO \\
	\vspace{-0.5cm}
  \{lei.wang, piotr.koniusz\}@data61.csiro.au \quad jun\_liu@sutd.edu.sg \\
	\vspace{-0.3cm}
}

\maketitle

\begin{abstract}
   In this paper, we propose a Few-shot Learning pipeline for 3D skeleton-based action recognition by Joint tEmporal and cAmera viewpoiNt alIgnmEnt (JEANIE). To factor out misalignment between query and support sequences of 3D body joints, we propose an advanced variant of Dynamic Time Warping which jointly models each smooth path between the query and support frames to achieve simultaneously the best alignment in the temporal and simulated camera viewpoint spaces for end-to-end learning under the limited few-shot training data. Sequences are encoded with a temporal block encoder based on Simple Spectral Graph Convolution, a lightweight linear Graph Neural Network backbone (we also include a setting with a transformer).  Finally, we propose a similarity-based loss which encourages the alignment of sequences of the same class while preventing the alignment of unrelated sequences. We demonstrate state-of-the-art results on NTU-60, NTU-120, Kinetics-skeleton and UWA3D Multiview Activity II.
\end{abstract}


\section{Introduction}
\label{sec:intro}

Action recognition (AR) has been studied for many decades~\cite{bobick2001, dollar2005, ke2007, laptev2008, bregonzio2009, Li2010, Shotton2011, Yang2012, Oreifej2013, Amor2016, Feichtenhofer_2017_CVPR,Carreira_2017_CVPR, lei_iccv_2019}, and it remains a vital topic in Computer Vision (CV) due to its numerous applications in the video surveillance, human-computer interaction, sports analysis, virtual reality and robotics. AR  pipelines~\cite{Tran_2015_ICCV,Feichtenhofer_2016_CVPR,Feichtenhofer_2017_CVPR,Carreira_2017_CVPR,lei_tip_2019,hosvd}  typically  perform action classification given the large amount of labeled training data. However, manually collecting and labeling videos or 3D skeleton-based sequences is laborious, and such pipelines need to be retrained or fine-tuned for new class concepts. The popular  AR networks include two-stream neural networks~\cite{Feichtenhofer_2016_CVPR, Feichtenhofer_2017_CVPR, Wang_2017_CVPR} and 3D convolutional networks (3D CNNs)~\cite{Tran_2015_ICCV, Carreira_2017_CVPR}, which  aggregate  frame-wise and temporal block representations, respectively. However, such networks indeed must be  trained on a large-scale dataset such as Kinetics \cite{Carreira_2017_CVPR} under a fixed set of training class concepts. 

Thus, there exists a growing interest in devising effective  Few-shot Learning (FSL) models for AR, termed Few-shot Action Recognition (FSAR), that rapidly adapt to  novel classes given a few training samples~\cite{mishra2018generative,xu2018dense,guo2018neural,dwivedi2019protogan,hongguang2020eccv, kaidi2020cvpr}. However, FSAR approaches that learn from videos are scarce. Due to the volumetric nature of videos and large intra-class variations,  it is hard to learn a reliable metric on the space of videos  given a few of clips per class. 

In contrast, FSL for the object category recognition has been studied now for many years 
\cite{miller_one_example,Li9596,NIPS2004_2576,BartU05,fei2006one,lake_oneshot} including 
a contemporary bulk of CNN-based FSL methods  
\cite{meta25,f4Matching,f1,f5Model-Agnostic,f8Relation,sosn}, which use meta-learning, prototype-based learning and feature representation learning. Just in 2020--2021 alone, the majority of new FSL methods focus on image classification~\cite{guo2020broader, nikita2020eccv, shuo2020eccv, moshe2020eccv, qinxuan2021wacv, nanyi2020accv, jiechao2020accv, kai2020cvpr, thomas2020cvpr, kaidi2020cvpr, luming2020cvpr}. In this paper, we build on the success of FSL with the goal of advancing recognition of articulated set of connected 3D body joints.

As shown by Johansson in his seminal experiment involving the moving lights display~\cite{johansson_lights}, a powerful alternative to video-based AR pipelines includes 3D skeleton-based AR pipelines, where  a human body is represented as an articulated set of connected 3D body joints that evolve in time~\cite{zatsiorsky_body}. Thanks to the Microsoft Kinect and advanced human pose estimation algorithms~\cite{Cao_2017_CVPR},  datasets for skeleton-based AR have become  accurate and widely available. The 3D skeleton-based AR  has  a rich history, including numerous shallow  approaches \cite{hussein_action,lv_3daction,parameswaran_viewinvariance,wu_actionlets,yang_eigenjoints,ofli_infjoints,vemulapalli_SE3} and  deep approaches \cite{7926607,stgcn2018aaai,Si_2019_CVPR,Li_2019_CVPR,Cheng_2020_CVPR,Liu_2020_CVPR} based on Long Short-Term Memory networks (LSTM) \cite{HochSchm97} or Graph Convolutional Networks (GCN) \cite{kipf2017semi}. 

With an exception of very recent models  \cite{liu_fsl_cvpr_2017,Liu_2019_NTURGBD120,2021dml,memmesheimer2021skeletondml}, FSAR approaches that learn from skeleton-based 3D body joints are  scarce. The above situation prevails despite AR from articulated sets of connected body joints, expressed as 3D coordinates, does offer a number of advantages over videos such as (i) the lack of the background clutter, (ii) the volume of data being several orders of magnitude smaller, and (iii) the 3D geometric manipulations of sequences being relatively friendly.  


\begin{figure*}[t]
\vspace{-0.3cm}
\includegraphics[width=18cm]{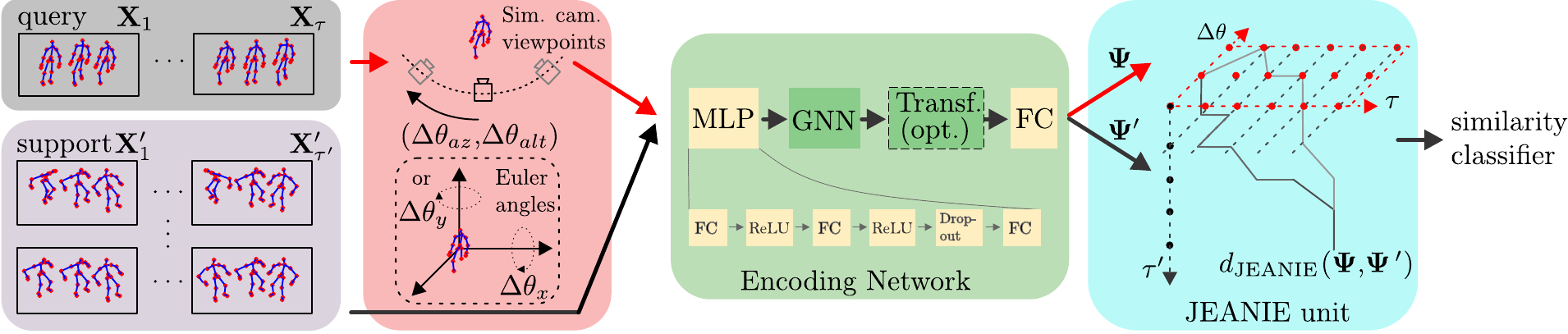}
%
%
%
%
\caption{Our 3D skeleton-based FSAR with JEANIE. Frames from a query sequence and a support sequence are split into short-term temporal blocks $\mX_1,\cdots,\mX_{\tau}$ and $\mX'_1,\cdots,\mX'_{\tau'}$ of length $M$ given stride $S$. Subsequently, we either generate (i) multiple rotations by $(\Delta\theta_x,\Delta\theta_y)$ of each query skeleton by either Euler angles (baseline approach) or (ii) simulated camera views (gray cameras) by camera shifts  $(\Delta\theta_{az},\Delta\theta_{alt})$  \wrt the assumed average camera location (black camera). We pass all skeletons via Encoding Network (with an optional transformer) to obtain feature tensors $\mPsi$ and $\mPsi'$, which are directed to JEANIE.
We note that the temporal-viewpoint alignment takes place in 4D space (we show a 3D case). While temporally-wise we start from the same $t\!=\!(1,1)$ and finish at $t\!=\!(\tau,\tau')$ (as in DTW), viewpoint-wise we need to start from every possible camera shift and finish at one of possible camera shifts. However, at each step, the path may move by no more than  $(\pm\!\Delta\theta_{az},\pm\!\Delta\theta_{alt})$.}
\label{fig:pipe}
\vspace{-0.3cm}
\end{figure*}

Motivated by the above observations, we propose a  FSAR  approach that learns on   skeleton-based 3D body joints via Joint tEmporal and cAmera  viewpoiNt alIgnmEnt (JEANIE). As FSL is based on learning similarity between support-query pairs, to achieve good matching of queries with support sequences representing the same action class, we propose to simultaneously model the optimal (i) temporal and (ii) viewpoint alignments. To this end, we build on  soft-DTW \cite{marco2017icml}, a differentiable variant of Dynamic Time Warping (DTW) \cite{marco2011icml}. Unlike soft-DTW, we exploit the projective camera geometry.  We assume that the best smooth path in DTW should simultaneously provide the best temporal and viewpoint alignment, as sequences that are being matched might have been captured under different camera viewpoints or subjects might have followed different trajectories. 

To obtain multiple skeleton representations under several viewpoints, we investigate rotating  skeletons (zero-centered by hip) by Euler angles \cite{eulera} \wrt $x$, $y$ and $z$ axes, and by generating skeleton locations given simulated camera positions, according to the algebra of stereo projections \cite{sterproj}.

As we model 
the viewpoint alignment, we notice that  view adaptive models are not an entirely new proposition in AR. View Adaptive Recurrent Neural Networks \cite{Zhang_2017_ICCV,8630687} is a classification model equipped with a view adaptation subnetwork that contains the rotation and translation switches within its RNN backbone, and the main LSTM-based network. Temporal Segment Network \cite{wang_2019_tpami} models long-range temporal structures with a new segment-based sampling and aggregation module. However, the above two AR classification pipelines  require a large number of training samples with variations of viewpoints and temporal shifts  to learn a robust model. The limitation of such models is evident 
when a network trained under a  fixed set of action classes has to be adapted to samples of novel classes.

In contrast, our FSAR  model learns end-to-end on skeleton-based 3D body joints in the powerful meta-learning regime to be able to  deal with joint temporal and viewpoint misalignment 
to recognize a handful of test samples of novel (previously unseen) category. 

Our pipeline consists of an MLP  which takes neighboring frames to form a temporal block. Firstly, we sample desired Euler rotations or simulated camera viewpoints, generate multiple skeleton views, and pass them to the MLP to get block-wise feature maps, next forwarded to a Graph Neural Network (GNN) \eg, GCN~\cite{kipf2017semi}, SGC~\cite{felix2019icml}, APPNP~\cite{johannes2019iclr} or S$^2$GC~\cite{hao2021iclr}, followed by an optional transformer~\cite{dosovitskiy2020image}, and an FC layer to obtain graph-based feature representations passed to our JEANIE and the similarity classifier. 

Finally, we note that JEANIE builds on the family of Reproducing Kernel Hilbert Space (RKHS) methods  \cite{Smola03kernelsand} which scale gracefully to FSAR problems which, by their setting, have to learn to match pairs of sequences  rather than learn a classifier of given action. In fact, JEANIE is an advanced variant inheriting from well-established principles of Optimal Transport \cite{ot}. JEANIE enjoys a transportation plan adapted to handle temporal and viewpoint alignment required by skeletal AR. 

\vspace{0.2cm}
Below are our contributions:
%
%
\renewcommand{\labelenumi}{\roman{enumi}.}
\vspace{-0.1cm}
\hspace{-1.0cm}
\begin{enumerate}[leftmargin=0.6cm]
\item We propose a Few-shot Action Recognition approach for learning on skeleton-based articulated 3D body joints via JEANIE, which performs the joint alignment of temporal blocks  and simulated viewpoint indexes of skeletons between support-query sequences to select 
the smoothest path without abrupt jumps in matching temporal locations and view indexes. How to warp temporal locations and simulated viewpoint indexes simultaneously is important in meta-learning, where samples of unseen classes have to be recognised.\vspace{-0.2cm}
\item To simulate different viewpoints of 
3D  skeleton  sequences, we consider rotating them  (1) by Euler angles within a specified range along $x$ and $y$ axes, or (2) to the  simulated camera locations based on the  algebra of stereo projection equations. These are normal operations in 3D geometry and AR but we use them as prerequisites for our joint alignment.
\vspace{-0.2cm}
\item We investigate several different GNN backbones (including transformer), as well as the optimal temporal size $\tau$ and stride for temporal blocks encoded by a simple 3-layer MLP unit before forwarding them to GNN.\vspace{-0.2cm}
\item We propose a simple  similarity-based loss encouraging the alignment of within-class sequences and preventing the alignment of between-class sequences, with a simple parameter to control the notion of similarity. 
\end{enumerate}

We achieve the state of the art  on 
 large-scale NTU-60 \cite{Shahroudy_2016_NTURGBD}, recent  NTU-120 \cite{Liu_2019_NTURGBD120} and Kinetics-skeleton~\cite{stgcn2018aaai} and  UWA3D Multiview Activity II \cite{Rahmani2016}.
%
%
The simultaneous alignment in joint temporal-viewpoint spaces has never been  proposed before in the meta-learning setting.





\section{Related Works}
\label{sec:rel}
Below, we describe video AR, 3D skeleton-based  AR,  FSL and FSAR approaches, 
and Graph Neural Networks. 

\vspace{0.05cm}
\noindent\textbf{AR (videos).} 
Many CNN- and LSTM-based AR pipelines \cite{cnn_basic_ar,cnn_lstm_ar,Feichtenhofer_2016_CVPR,cnn3d_ar,spattemp_filters,spat_temp_resnet,long_term_ar} exist with early ones extracting per-frame representations for average  \cite{cnn_basic_ar} 
or LSTM-based pooling \cite{cnn_lstm_ar}. Two-stream networks \cite{Feichtenhofer_2016_CVPR} 
used RGB and the optical flow network streams. Spatio-temporal 3D CNN filters become an AR standard 
\cite{cnn3d_ar,spattemp_filters,spat_temp_resnet,long_term_ar}. The above AR methods rely on big datasets with predefined categories. 

\vspace{0.05cm}
\noindent\textbf{AR (3D skeletons).} Since early works on   
3D skeleton AR  \cite{zatsiorsky_body}, several approaches modeled 3D joints  dynamics 
\cite{hussein_action,lv_3daction},  orientations of joints 
\cite{parameswaran_viewinvariance} 
and  relative 3D joint locations \cite{wu_actionlets,yang_eigenjoints}. Others used  connected body segments \cite{yacoob_activities,ohn_hog2,ofli_infjoints}, special {\em SE3} group \cite{vemulapalli_SE3}, temporal hierarchy of coefficients \cite{hussein_action}, pair-wise positions \cite{wu_actionlets}, spatio-temporal correlations \cite{lei_kermats,cavazza_kercov,lei_sice},  time-series kernels \cite{gaidon_timekern} and tensors \cite{tensor_ar_old,hosvd} described in survey papers \cite{presti20153d,lei_tip_2019}.

Currently, the best  3D skeletons  AR pipelines build on  LSTM \cite{HochSchm97} or GCNs \cite{kipf2017semi} \eg, 
an  LSTM  with geometric relational features \cite{7926607}, spatio-temporal GCN \cite{stgcn2018aaai},  an attention enhanced graph convolutional LSTM network \cite{Si_2019_CVPR},   a so-called A-links inference model \cite{Li_2019_CVPR},  shift graph operations and lightweight point-wise convolutions \cite{Cheng_2020_CVPR} and multi-scale aggregation node disentangling scheme 
\cite{Liu_2020_CVPR}. However, the above AR methods 
rely on large scale datasets, and are difficult to be adapted 
 to new class concepts, unlike FSAR approaches detailed below.


\vspace{0.05cm}
\noindent\textbf{FSAR (videos).} 
Approaches \cite{mishra2018generative,guo2018neural,xu2018dense} use a generative model, graph matching on 3D coordinates and dilated networks, 
respectively. Approach \cite{Zhu_2018_ECCV} uses a  compound memory network. 
ProtoGAN \cite{dwivedi2019protogan}  generates action prototypes. Recent FSAR approach \cite{hongguang2020eccv} uses permutation-invariant attention and second-order aggregation of temporal video blocks, whereas approach   \cite{kaidi2020cvpr} proposes a modified temporal alignment for query-support pairs via DTW. 

\vspace{0.05cm}
\noindent\textbf{FSAR (3D skeletons).} 
Very few FSAR approaches work with 3D skeletons 
\cite{liu_fsl_cvpr_2017,Liu_2019_NTURGBD120,2021dml,memmesheimer2021skeletondml}. Global Context-Aware Attention LSTM \cite{liu_fsl_cvpr_2017} selectively focuses on the informative joints. 
Action-Part Semantic Relevance-aware (APSR) framework \cite{Liu_2019_NTURGBD120}  uses  the  semantic relevance between each body part and each action class at the  distributed  word  embedding  level. 
Signal Level Deep Metric Learning \cite{2021dml}  and Skeleton-DML \cite{memmesheimer2021skeletondml} one-shot FSL approaches  encode  signals  into  images,  extract  features  using  a  deep  residual  CNN and apply multi-similarity miner losses. 
In contrast, our approach uses temporal blocks of  3D body joints of skeletons encoded by GNNs under multiple viewpoints of skeletons 
to simultaneously perform temporal and viewpoint-wise  alignment of query-support in the meta-learning regime.

\vspace{0.05cm}
\noindent\textbf{Graph Neural Networks.} GNNs have been very successful in the skeleton-based AR, as already discussed in previous paragraphs. Below, we provide a brief background on generic GNNs, on which we build in this paper due to their excellent ability to represent graph structured data such as interconnected body joints. GCN \cite{kipf2017semi} applies graph convolution in the spectral domain, and enjoys the depth-efficiency when stacking multiple layers due to non-linearities. However, depth-efficiency costs speed due to backpropagation through consecutive layers. In contrast, a very recent family of so-called spectral filters do not require depth-efficiency but apply filters based on heat  diffusion to the graph Laplacian. As a result, they are fast linear models as learnable weights act on filtered node representations. SGC~\cite{felix2019icml}, APPNP~\cite{johannes2019iclr} and S$^2$GC~\cite{hao2021iclr} are  three methods from this family which we investigate for the backbone. 



\vspace{0.05cm}
\noindent\textbf{Multi-view AR.} 
%
Skeleton-based AR represent humans under various camera viewpoints. Some multi-view AR methods explore complementary modalities to bridge the gap between different views \eg, one modality may miss at a given view \cite{6719542}. 
Multi-view AR gains some traction \cite{lei_tip_2019, Zhang_2017_ICCV} due to multi-modal sensors. An audio-visual tracker \cite{6719542} can recover in absence of one modality. A Generative Multi-View Action Recognition framework  \cite{Wang_2019_ICCV}  integrates complementary information from  RGB and depth sensors by View Correlation Discovery Network. Others exploit multiple views of the  action \cite{Shahroudy_2016_NTURGBD, Liu_2019_NTURGBD120,8630687,Wang_2019_ICCV} to overcome the viewpoint variations but are dedicated to AR recognition with a fixed set of classes and large training datasets.

In contrast, our  JEANIE learns to perform jointly the  temporal and simulated viewpoint alignment 
 in an end-to-end meta-learning setting. This is a novel paradigm  based on similarity learning of support-query pairs rather than learning class concepts as in classical AR.

\section{Prerequisites}
\label{sec:back}
Below, we present our notations, a necessary background on Euler angles, the algebra of stereo projection, GNNs used in this work and the formulation of soft-DTW.

\vspace{0.05cm}
\noindent\textbf{Notations.}  $\idx{K}$ stands for the index set $\{1,2,\cdots,K\}$. Concatenation of $\alpha_i$ is denoted by $[\alpha_i]_{i\in\idx{I}}$, whereas $\mX_{:,i}$ means we extract/access column $i$ of matrix $\mD$. Calligraphic mathcal fonts denote tensors (\eg, $\tD$), capitalized bold symbols are matrices (\eg, $\mD$), lowercase bold symbols  are vectors (\eg, $\vpsi$), and regular fonts denote scalars. 


\vspace{0.05cm}
\noindent{{\bf Euler angles \cite{eulera}}}  are defined as successive planar rotation angles around $x$, $y$, and $z$ axes.  For 3D coordinates, we have  the following rotation matrices ${\bf R}_x$, ${\bf R}_y$ and ${\bf R}_z$:
\vspace{-0.2cm}
\begin{align}
\left[\arraycolsep=1.4pt\def\arraystretch{0.5}\begin{array}{ccc}
1 & 0 & 0\\
0 & \text{cos}\theta_x & \text{sin}\theta_x\\
0 & -\text{sin}\theta_x & \text{cos}\theta_x
\end{array} 
\right ],
\left[\arraycolsep=1.4pt\def\arraystretch{0.5}\begin{array}{ccc}
\text{cos}\theta_y & 0 & -\text{sin}\theta_y\\
0 & 1 & 0\\
\text{sin}\theta_y & 0 & \text{cos}\theta_y
\end{array} 
\right],
\left[\arraycolsep=1.4pt\def\arraystretch{0.5} \begin{array}{ccc}
\text{cos}\theta_z & \text{sin}\theta_z &  0\\
-\text{sin}\theta_z & \text{cos}\theta_z & 0\\
0 & 0 & 1
\end{array} 
\right]
\end{align}

\vspace{-0.2cm}
\noindent As the resulting composite rotation matrix  depends on the order of  rotation axes \ie, $\mathbf{R}_x\mathbf{R}_y\mathbf{R}_z\!\neq\!\mathbf{R}_z\mathbf{R}_y\mathbf{R}_x$, we also investigate    the algebra of stereo projection presented next.

\comment{ Similarly, rotating along $x$ and $z$, the rotation matrix are $R_x\!=\! 
\left[ \begin{array}{ccc}
1 & 0 & 0\\
0 & \text{cos}\theta_x & \text{sin}\theta_x\\
0 & -\text{sin}\theta_x & \text{cos}\theta_x
\end{array} 
\right ]$ and $R_z= 
\left[ \begin{array}{ccc}
\text{cos}\theta_z & \text{sin}\theta_z &  0\\
-\text{sin}\theta_z & \text{cos}\theta_z & 0\\
0 & 0 & 1
\end{array} 
\right ]$ respectively.

Based on the above rotation matrices, we can get the rotation matrix for rotating along $y$-$x$-$z$ is:
\begin{tiny}
\begin{equation}
{\bf R}_{yxz}= 
\left[ \begin{array}{ccc}
\text{cos}\theta_y\text{cos}\theta_z-\text{sin}\theta_x\text{sin}\theta_y\text{sin}\theta_z & \text{cos}\theta_y\text{sin}\theta_z+\text{sin}\theta_x\text{sin}\theta_y\text{cos}\theta_z & -\text{sin}\theta_y\text{cos}\theta_x\\
-\text{cos}\theta_x\text{sin}\theta_z & \text{cos}\theta_x\text{cos}\theta_z & \text{sin}\theta_x\\
\text{sin}\theta_y\text{cos}\theta_z+\text{sin}\theta_x\text{cos}\theta_y\text{sin}\theta_z & \text{sin}\theta_y\text{sin}\theta_z-\text{sin}\theta_x\text{cos}\theta_y\text{cos}\theta_z & \text{cos}\theta_x\text{cos}\theta_y
\end{array} 
\right ]
\end{equation}
\end{tiny}}

\vspace{0.05cm}
\noindent{{\bf Stereo projections}} \cite{sterproj}. \comment{Let left/right cameras, with ${\bf R}_l$ and ${\bf R}_r$ rotation matrices and ${\bf T}_l$ and ${\bf T}_r$ translation matrices, capture a point ${\bf p}$ in the real world. We want to find a link between the point in the left and the right camera, that is, ${\bf p}_l$ and ${\bf p}_r$. 
We have ${\bf p}_l\!=\!{\bf R}_l{\bf p}\!+\!{\bf T}_l$ and ${\bf p}_r\!=\!{\bf R}_r{\bf p}\!+\!{\bf T}_r$. One can write 
\vspace{-0.2cm}
\begin{equation}{\bf p}_l = {\bf R}^T({\bf p}_r-{\bf T})
\end{equation}
to get the relationship between two views in two cameras, where 
\vspace{-0.3cm}
\begin{equation}
\begin{split}
{\bf R} \!=\! {\bf R}_r({\bf R}_l) {\bf T},\\
{\bf T} \!=\! {\bf T}_r - {\bf R}{\bf T}_l.
\end{split}
\end{equation}

\noindent Thus, one just needs to know the position/orientation of the left camera relative to the right camera. We imagine locations of 3D body joints by offsetting the imaginary left camera \wrt  the assumed location of actual (right) camera. }
%
Suppose we have a rotation matrix ${\bf R}$ and a translation vector ${\bf t}\!=\![t_x, t_y, t_z]^T$ between left/ right cameras (imagine some non-existent stereo camera). Let ${\bf M}_l$ and ${\bf M}_r$ be  the intrinsic matrices of the left/right cameras. 
Let ${\bf p}_l$ and ${\bf p}_r$ be coordinates of the left/right camera. As the origin of the right camera in the left camera coordinates is ${\bf t}$,  we have: ${\bf p}_r\!=\!{\bf R}({\bf p}_l\!-\!{\bf t})$ and  
$({\bf p}_l\!-\!{\bf t})^T\!=\!({\bf R}^{T}{\bf p}_r)^T$. 
The plane  (polar surface) formed by all points passing through ${\bf t}$ can be expressed by $({\bf p}_l\!-\!{\bf t})^T({\bf p}_l\!\times\!{\bf t})\!=\!0$. Then, ${\bf p}_l\!\times \!{\bf t}\!=\!{\bf S}{\bf p}_l$ where ${\bf S}\!=\! \left[\arraycolsep=1.4pt\def\arraystretch{0.5}\begin{array}{ccc}
0 & -t_z & t_y \\
t_z & 0 & -t_x\\
-t_y & t_x & 0
\end{array} 
\right]$. 
Based on the above equations, we obtain ${{\bf p}_r}^T{\bf R}{\bf S}{\bf p}_l\!=\!0$, and note that ${\bf R}{\bf S}\!=\!{\bf E}$ is the Essential Matrix, and ${{\bf p}^T_r}{\bf E} {\bf p}_l\!=\!0$ describes the relationship for the same physical point under the left and right camera coordinate system. 
As ${\bf E}$ has no internal inf. about the camera, and ${\bf E}$ is based on the camera coordinates, we  use a fundamental matrix {\bf F} that describes the relationship for the same physical point under the camera pixel coordinate system. 
The relationship between the pixel  and camera coordinates is: ${\bf p}^*\!=\!{\bf M}{\bf p}'$ and ${{\bf p}'_r}^T{\bf E} {\bf p}'_l\!=\!0$.

Now, suppose the pixel coordinates of ${\bf p}'_l$ and ${\bf p}'_r$ in the pixel coordinate system are ${\bf p}^*_{l}$ and ${\bf p}^*_{r}$, then we can write ${{\bf p}^*_{r}}^T({\bf M}_r^{-1})^T{\bf E}{\bf M}_l^{-1}{\bf p}^*_{l}\!=\!0$, where  ${\bf F}\!=\!({\bf M}_r^{-1})^T{\bf E}{\bf M}_l^{-1}$ is the fundamental matrix. Thus, the relationship for the same point in the pixel coordinate system of the left/right camera is:
\vspace{-0.2cm}
\begin{equation}
{{\bf p}^*_{r}}^{T}{\bf F}{\bf p}^*_{l}\!=\!0.
\label{eq:f_matrix}
\end{equation}

\vspace{-0.0cm}
\noindent We treat 3D body joint coordinates as ${\bf p}^*_{l}$. Given ${\bf F}$ (estimation of ${\bf F}$ is  explained later), we obtain their coordinates ${\bf p}^*_{r}$ in the new view.

\vspace{0.05cm}
\noindent{\bf GNN notations.} 
Firstly, let $G\!=\!(\bf{V}, {\bf E})$ be a graph with the vertex set $\bf{V}$  with nodes $\{v_1, \cdots, v_n\}$, and ${\bf E}$ are edges of the graph. Let ${\bf A}$ and ${\bf D}$ be the adjacency and diagonal degree matrix, respectively. Let $\tilde{\bf A}\!=\!{\bf A}\!+\!{\bf I}$ be the adjacency matrix with self-loops (identity matrix) with the corresponding diagonal degree matrix $\tilde{\bf D}$ such that $\tilde{D}_{ii}\!=\!\sum_j ({\bf A}^{ij}\!+\! {\bf I}^{ij})$. 
Let ${\bf S}\!=\!\tilde{\bf D}^{-\frac{1}{2}} \tilde{\bf A}\tilde{\bf D}^{-\frac{1}{2}}$ be the normalized adjacency matrix with added self-loops. For the $l$-th layer, we use ${\bf \Theta}^{(l)}$ to denote the learnt weight matrix, and ${\bf \Phi}$ to denote the outputs from the graph networks. Below, we list backbones used by us.

\vspace{0.05cm}
\noindent{\bf GCN}~\cite{kipf2017semi}. GCNs learn the feature representations for the features $\mathbf{x}_i$ of each node over multiple layers. For the $l$-th layer, we denote the input  by  ${\bf H}^{(l-1)}$ and the output by ${\bf H}^{(l)}$. Let the input (initial) node representations be ${\bf H}^{(0)}\!=\! {\bf X}$. 
For an $L$-layer GCN, the output representations are given by:
\begin{equation}
    {\bf \Phi_\text{GCN}}\!=\!{\bf S}{\bf H}^{(L-1)}{\bf \Theta}^{(L)} \text{ where } {\bf H}^{(l)}\!\!=\!\text{ReLU}({\bf S}{\bf H}^{(l-1)}{\bf \Theta}^{(l)}).
\end{equation}

\vspace{0.05cm}
\noindent{{\bf APPNP}}~\cite{johannes2019iclr}. The Personalized Propagation of Neural Predictions (PPNP) and its fast approximation, APPNP, are based on the  personalized PageRank. Let  ${\bf H}^{(0)}\!=\!f_{\bf \Theta}(X)$ be the input to APPNP, where $f_{\bf \Theta}$ can be an MLP with parameters  ${\bf \Theta}$. Let the output of the $l$-th layer be ${\bf H}^{(l)}\!=\! (1-\alpha){\bf S}{\bf H}^{(l-1)}\!+\!\alpha{\bf H}^{(0)}$, where $\alpha$ is the teleport (or restart) probability  in range $(0, 1]$. For an $L$-layer APPNP, we have:
\vspace{-0.2cm}
\begin{equation}
    {\bf \Phi_\text{APPNP}}\!=\!{(1\!-\!\alpha)\bf S}{\bf H}^L\!+\!\alpha{\bf H^{(0)}}.
\end{equation}
\vspace{-0.4cm}


\vspace{0.05cm}
\noindent{{\bf SGC}}~\cite{felix2019icml} { \&} {{\bf S$^2$GC}}~\cite{hao2021iclr}. 
SGC captures the  $L$-hops neighbourhood in the graph by the $L$-th power of the transition matrix used as a spectral filter. For an $L$-layer SGC, 
we obtain:
\vspace{-0.2cm}
\begin{equation}
    {\bf \Phi_\text{SGC}}\!=\!{\bf S}^L{\bf X}{\bf \Theta}.
\end{equation}
\vspace{-0.5cm}

Based on a modified Markov Diffusion Kernel, Simple Spectral Graph Convolution (S$^2$GC) is the summation over $l$-hops, $l\!=\!1,\cdots,L$.  The output of S$^2$GC is: 
\vspace{-0.2cm}
\begin{equation}
    {\bf \Phi_\text{S$^2$GC}} \!=\!  \frac{1}{L}\sum_{l=1}^{L}((1\!-\!\alpha){\bf S}^l{\bf X}\!+\!\alpha{\bf X}){\bf \Theta}.
\end{equation}
\vspace{-0.2cm}


\vspace{0.05cm}
\noindent{\bf Soft-DTW \cite{marco2011icml,marco2017icml}.} Dynamic Time Warping can be seen as a specialized  case of the Wasserstein metric, under specific transportation plan. Soft-DTW is defined as:
\begin{align}
& d_{\text{DTW}}(\mPsi,\mPsi')\!=\!\softming\limits_{\mA\in\tAnb_{\tau,\tau'}}\left\langle\mA,\mD(\mPsi,\mPsi')\right\rangle,\\
& \text{ where } \softming(\valpha)\!=\!-\gamma\!\log\sum_i\exp(-\alpha_i/\gamma).
\end{align}
The binary  $\mA\!\in\!\tAnb_{\tau,\tau'}$ denotes a path within the transportation plan $\tAnb_{\tau,\tau'}$ which depends on lengths $\tau$ and $\tau'$ of sequences $\mPsi\!\equiv\![\vpsi_1,\cdots,\vpsi_\tau]\!\in\!\mbr{d'\times\tau}$, $\mPsi'\!\equiv\![{\vpsi'}_1,\cdots,{\vpsi'}_{\tau'}]\!\in\!\mbr{d'\times\tau'}$ and  $\mD\!\in\!\mbrp{\tau\times\tau'}\!\!\equiv\![d_{\text{base}}(\vpsi_m,\vpsi'_n)]_{(m,n)\in\idx{\tau}\times\idx{\tau'}}$, the matrix of distances, is evaluated for $\tau\!\times\!\tau'$ frame representations according to some base distance $d_{\text{base}}(\cdot,\cdot)$ \ie, the Euclidean or the RBF-induced distance. In what follows, we make use of principles of soft-DTW. However, we design a joint alignment between temporal skeleton sequences and simulated skeleton viewpoints, an entirely novel proposal.

\section{Approach}
\label{sec:appr}

\begin{figure}[t]
\vspace{-0.3cm}
\centering
\begin{subfigure}[t]{0.49\linewidth}
\includegraphics[width=4.0cm]{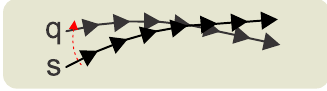}
\caption{\label{fig:ttra}}
\end{subfigure}
\begin{subfigure}[t]{0.49\linewidth}
\includegraphics[width=4.0cm]{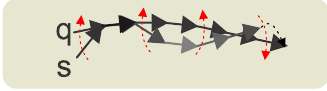}
\caption{\label{fig:ttrb}}
\end{subfigure}
\begin{subfigure}[t]{0.49\linewidth}
\includegraphics[width=4.0cm]{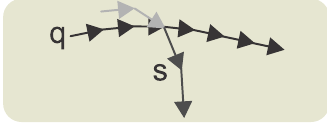}
\caption{\label{fig:ttrc}}
\end{subfigure}
\begin{subfigure}[t]{0.49\linewidth}
\includegraphics[width=4.0cm]{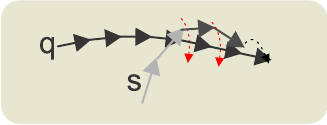}
\caption{\label{fig:ttrd}}
\end{subfigure}
\vspace{-0.3cm}
\caption{Different matching scenarios  of query $q$ and support $s$.}
\label{fig:s_q_match}
\vspace{-0.3cm}
\end{figure}


Our goal is to find a smooth joint viewpoint-temporal alignment between pairs of sequences (query and support) and minimize or maximize the matching distance $d_\text{JEANIE}$ (end-to-end setting) for same or different support-query labels, respectively. 
Fig.~\ref{fig:ttra} shows that sometimes matching of $q$ and $s$ may be as easy as rotating one trajectory onto another. However, in fact $q$ and $s$ may be of different classes and so such an alignment is superficial. 
%
Fig.~\ref{fig:ttrb} shows the case where  the $q$ and  $s$ pair (same class) does not perfectly match but smoothly rotating individual blocks of $s$ (along red arrows) achieves a good match (the grey arrows indicating the partial feature similarity will turn dark indicating a very good feature match). 
Fig.~\ref{fig:ttrc} shows that naively matching two trajectories by their angle may match regions in which features of $s$ (light gray) do not match features of $q$. This bad outcome is likely for traditional viewpoint matching. 
Fig.~\ref{fig:ttrd} shows that JEANIE aligns and warps $s$ to $q$ based on the feature similarity (grey arrows are not similar to dark ones so JEANIE aligns along more similar feature pairs which can be smoothly warped in viewpoint and temporal sense so as to further improve feature matching of $q$ and $s$ in aligned regions).


Below, we detail our pipeline shown in Figure \ref{fig:pipe}, and explain the proposed JEANIE and our loss function.

\vspace{0.05cm}
\noindent{\bf Encoding Network (EN).}  
We start by generating $K\!\times\!K'$ Euler rotations or $K\!\times\!K'$ simulated camera views (moved gradually from the estimated camera location) of query skeletons. 
Our EN contains a simple 3-layer MLP unit (FC, ReLU, FC, ReLU, Dropout, FC), GNN, optional  Transformer~\cite{dosovitskiy2020image} and FC. 
The MLP unit takes $M$ neighboring frames, each with $J$ 3D skeleton body joints, forming one temporal block. In total, depending on stride $S$, we obtain some $\tau$ temporal blocks which capture the short temporal dependency, whereas the long temporal dependency will be modeled with our JEANIE. Each temporal block is encoded by the MLP into a $d\!\times\!J$ dimensional feature map.  Subsequently,  query feature maps of size $K\!\times\!K'\!\times\!\tau$ and support feature maps of size $\tau'$  are  forwarded to a GNN, optional Transformer (similar to ViT~\cite{dosovitskiy2020image}, instead of using image patches, we feed each body joint encoded by GNN into the transformer), and  an FC layer, which returns $\mPsi\!\in\!\mbr{d'\times K\times K'\times\tau}$ query feature maps and $\mPsi'\!\in\!\mbr{d'\times\tau'}\!$ support feature maps. 
Such encoded feature maps are  passed to JEANIE and the similarity classifier. 

Let support  maps  $\mPsi'\!$ be 
$[f(\boldsymbol{X}'_1;\mathcal{F}),\cdots,f(\boldsymbol{X}'_{\tau'};\mathcal{F})]\!\in\!\mbr{d'\times\tau'}$
and query maps $\mPsi$ be 
$[f(\boldsymbol{X}_1;\mathcal{F}),\cdots,f(\boldsymbol{X}_\tau;\mathcal{F})]\!\in\!\mbr{d'\times K\times K'\times\tau}$, for 
query and support frames per block $\mX,\mX'\!\in\!\mbr{3\times J\times M}$. Moreover, we define {\fontsize{9}{9}\selectfont$f(\mX; \mathcal{F})\!=\!\text{FC}(\text{Transf}(\text{GNN}(\text{MLP}(\mX; \mathcal{F}_{MLP}); \mathcal{F}_{GNN}); \mathcal{F}_{Transf}); \mathcal{F}_{FC})$},  $\mathcal{F}\!\equiv\![\mathcal{F}_{MLP},\mathcal{F}_{GNN},\mathcal{F}_{Transf},\mathcal{F}_{FC}]$ is the  set of  parameters of EN (note optional Transformer \cite{dosovitskiy2020image}). As GNN, we try GCN~\cite{kipf2017semi}, SGC~\cite{felix2019icml}, APPNP~\cite{johannes2019iclr} or S$^2$GC~\cite{hao2021iclr}.

\vspace{0.1cm}
\noindent{\bf JEANIE.}  Matching query-support pairs requires temporal alignment due to potential offset in locations of discriminative parts of actions, and due to potentially different dynamics/speed of actions taking place. The same concerns the direction of the dominant action trajectory \wrt the camera. Thus, JEANIE, our advanced soft-DTW, has the  transportation plan  $\tAnb'\!\equiv\!\tAnb_{\tau,\tau', K, K'}$, where  apart from temporal block counts $\tau$ and $\tau'$, for query sequences, we have possible $\eta_{az}$ left and $\eta_{az}$ right steps from the initial camera azimuth, and $\eta_{alt}$ up and $\eta_{alt}$ down  steps from the initial camera altitude. Thus, $K\!=\!2\eta_{az}\!+\!1$, $K'\!=\!2\eta_{alt}\!+\!1$.  For the variant with Euler angles, we simply have $\tAnb''\!\equiv\!\tAnb_{\tau,\tau', K, K'}$ where $K\!=\!2\eta_{x}\!+\!1$, $K'\!=\!2\eta_{y}\!+\!1$ instead. Then, JEANIE is given as:
\vspace{-0.2cm}
\begin{equation}
d_{\text{JEANIE}}(\mPsi,\mPsi')\!=\!\softming\limits_{\mA\in\tAnb'}\left\langle\mA,\tD(\mPsi,\mPsi')\right\rangle,
\label{eq:d_jeanie}
\vspace{-0.3cm}
\end{equation}

\noindent
where  $\tD\!\in\!\mbrp{K\times\!K'\!\times\tau\times\tau'}\!\!\equiv\![d_{\text{base}}(\vpsi_{m,k,k'},\vpsi'_n)]_{\substack{ (m,n)\in\idx{\tau}\!\times\!\idx{\tau'}\\(k,k')\in\idx{K}\!\times\!\idx{K'}}},\!\!\!\!\!\!$
and tensor $\tD$ contains distances. 
Algorithm \ref{code:JEANIE} illustrates JEANIE. For brevity, let us tackle the camera viewpoint alignment  in a single space \eg, for some shifting steps $-\eta,\cdots,\eta$, each with  size  $\Delta\theta_{az}$. The maximum viewpoint change from block to block is $\iota$-max shift (smoothness). As we have no way to know the initial optimal camera shift, we create all possible origins of shifts  $\Delta n\!\in\!\{-\eta, \cdots, \eta\}$ (think of $\Delta n$ as the selector of shift origin). Subsequently, a  phase related to DTW (temporal-viewpoint alignment) takes place. Finally, we choose the path with the smallest distance  over all possible shift-starts and shift-ends.

\algblock{while}{endwhile}
\algblock[TryCatchFinally]{try}{endtry}
\algcblockdefx[TryCatchFinally]{TryCatchFinally}{catch}{endtry}
	[1]{\textbf{except}#1}{}
\algcblockdefx[TryCatchFinally]{TryCatchFinally}{elsee}{endtry}
	[1]{\textbf{else}#1}{}
\algtext*{endwhile}
\algtext*{endtry}

\algblock{for}{endfor}
\algtext*{endfor}

\algblockdefx{ifff}{endifff}
	[1]{\textbf{if}#1}{}
\algtext*{endifff}

\algblockdefx{elseee}{endelseee}
	[1]{\textbf{else}#1}{}
\algtext*{endelseee}

\begin{algorithm}[tbp!]
\caption{Joint tEmporal and cAmera viewpoiNt alIgnmEnt (JEANIE).}
\label{code:JEANIE}
{\bf Input} (forward pass): $\mPsi, \mPsi'$, $\gamma\!>\!0$, $d_{base}(\cdot,\cdot)$, $\iota$-max shift.
\begin{algorithmic}[1]
\State{$r_{:,:,:,:}\!=\!\infty, r_{0,0,0,0}\!=\!0$}
\for{ $\Delta n,n\!\in\!\{-\eta,\cdots,\eta\}$:}
\ifff{ $n\!-\!\Delta n\in\{-\eta,\cdots,\eta\}$:}
\for{ $t\!\in\!\idx{\tau}$:}
\for{ $t'\!\in\!\idx{\tau'}$:}
	\State{$\!\!\!\!\!\!\!\!\!\!\!\!\!\!\!\!\!\!\!\!\!\!\!\!\!\!\!\!\substack{r_{\Delta n,n,t,t'}=d_{base}(\vpsi_{n\!-\!\Delta n,t}, \vpsi'_{t'})+\\
	\qquad\qquad\softming\left([r_{\Delta n, n\!-\!i,t\!-\!j,t'\!-\!k}]_{\!\!\!\!\!\!\!\!\!\!\!\!\!\!\!\!\!\!\!\substack{(i,j,k)\in\Pi,\\\!\!\!\!\!\!\!\!\!\!\!\!\!\!\!\!\!\!\!\!\!\!\!\!\!\Pi\equiv\{0,\cdots,\iota\}\times\{0,1\}\times\{0,1\}\setminus\{(0,0,0)\}}}\right)
	}$}
\endfor
\endfor
\endifff
\endfor
\end{algorithmic}
{\bf Output:} $\softming\left([r_{\Delta n,n,\tau,\tau'}]_{\Delta n,n\in\{-\eta,\cdots,\eta\}}\right)$
\end{algorithm}

\vspace{0.05cm}
\noindent{\bf Loss Function.} For the $N$-way $Z$-shot problem, 
we have   one query feature map  and $N\!\times\!Z$ support feature maps per episode. We form a mini-batch containing $B$ episodes. Thus, we have query feature maps $\{\mPsi_b\}_{b\in\idx{B}}$ and support feature maps $\{\mPsi'_{b,n,z}\}_{b\in\idx{B},n\in\idx{N},z\in\idx{Z}}$. Moreover,  $\mPsi_b$ and $\mPsi'_{b,1,:}$ share the same class, one of $N$ classes drawn per episode, forming the  subset $C^{\ddagger} \equiv \{c_1,\cdots,c_N \} \subset \mathcal{I}_C \equiv \mathcal{C}$. To be precise, labels  $y(\mPsi_b)\!=\!y(\mPsi'_{b,1,z}), \forall b\!\in\!\idx{B}, z\!\in\!\idx{Z}$ while $y(\mPsi_b)\!\neq\!y(\mPsi'_{b,n,z}), \forall b\!\in\!\idx{B},n\!\in\!\idx{N}\!\setminus\!\{1\},  z\!\in\!\idx{Z}$. Moreover, in most cases, $y(\mPsi_b)\!\neq\!y(\mPsi_{b'})$ if $b\!\neq\!b'$ and $b,b'\!\in\!\idx{B}$. However, selection of $C^{\ddagger}$ per episode is done randomly. For the $N$-way $Z$-shot protocol, we minimize:
\vspace{-0.0cm}
\begin{align}
 & \!\!\!\!\!l(\vd^{+}\!,\vd^{-})\!=\!\left(\mu(\vd^{+})\!-\!\detach(\mu(\topminb(\vd^{+})))\right)^2\label{eq:pos}\\
 & \!\!\qquad\,+\!\left(\mu(\vd^{-})\!-\!\detach(\mu(\topmaxbb(\vd^{-})))\right)^2\!,\label{eq:neg}\\
 & \quad\;\text{ where }\quad\;\vd^{+}\!=\![d_{JEANIE}(\mPsi_{b},\mPsi'_{b,1,z})]_{\substack{b\in\idx{B}\\z\in\idx{Z}}}\nonumber\\
 & \quad\;\text{ and }\quad\;\vd^{-}\!=\![d_{JEANIE}(\mPsi_{b},\mPsi'_{b,n,z})]_{\!\!\!\!\!\substack{b\in\idx{B},\\n\in\idx{N}\!\setminus\!\{1\}\!, z\in\idx{Z}}},\nonumber
\end{align}
%
%
where $\vd^+$ is a set of within-class distances for the mini-batch of size $B$ given $N$-way $Z$-shot learning protocol. By analogy,  $\vd^-$ is a set of between-class distances. Function $\mu(\cdot)$ is simply the mean over coefficients of the input vector, $\detach(\cdot)$ detaches the graph during the backpropagation step, whereas $\topminb(\cdot)$ and $\topmaxbb(\cdot)$ return $\beta$ smallest and $NZ\beta$ largest coefficients from the input vectors, respectively. Thus, Eq. \eqref{eq:pos} promotes the within-class similarity while Eq. \eqref{eq:neg} reduces the between-class similarity. Integer $\beta\!\geq\!0$ controls the focus on difficult examples \eg, $\beta\!=\!1$  encourages all within-class distances in Eq.  \eqref{eq:pos} to be close to the positive target ($\mu(\topminb(\cdot))$), the smallest observed within-class distance in the mini-batch. If $\beta\!>\!1$, this means we relax our positive target.  By analogy, if $\beta\!=\!1$, we encourage all between-class distances in Eq.  \eqref{eq:neg} to approach the negative target ($\mu(\topmaxbb(\cdot))$), the average over the largest $NZ$ between-class distances. If $\beta\!>\!1$, the negative target is relaxed.




\section{Experiments}

We provide the training details in Sec.~\ref{network_train} of \textbf{Suppl. Material}.
Below, we describe the datasets and evaluation protocols on which we  validate our FSAR with JEANIE.


\vspace{0.05cm}
\noindent\textbf{Datasets.}  Section \ref{app:ds} and Table \ref{datasets} of \textbf{Suppl. Material} contain details of datasets described below.

\comment{
\vspace{0.05cm}
\noindent{\em{MSRAction3D}}~\cite{Li2010} is an older AR datasets captured with the Kinect depth camera. It contains 20 human sport-related activities such as {\it jogging}, {\it golf swing} and {\it side boxing}. 

\vspace{0.05cm}
\noindent{{\em 3D Action Pairs}}~\cite{Oreifej2013} contains 6 selected pairs of actions that have very similar motion trajectories \eg, {\it put on a hat} and {\it take off a hat}, {\it pick up a box} and {\it put down a box}, \etc. 

\vspace{0.05cm}
\noindent{{\em UWA3D Activity}}~\cite{RahmaniHOPC2014} has 30 actions performed by 10 people of various height at different speeds in cluttered scenes. 
}

\vspace{0.05cm}
\noindent{{\em UWA3D Multiview Activity II}}~\cite{Rahmani2016} contains 30 actions performed by 9 people in a cluttered environment. In this dataset, the Kinect camera was moved to different positions to capture the actions from 4 different views: front view ($V_1$), left view ($V_2$), right view ($V_3$), and top view ($V_4$). 

\vspace{0.05cm}
\noindent{{\em NTU RGB+D (NTU-60)}}~\cite{Shahroudy_2016_NTURGBD} contains 56,880 video sequences and over 4 million frames. 
This dataset has variable sequence lengths  and exhibits high intra-class variations. 

\vspace{0.05cm}
\noindent{{\em NTU RGB+D 120 (NTU-120)}}~\cite{Liu_2019_NTURGBD120}, an extension of NTU-60, contains 120 action classes (daily/health-related), and 114,480 RGB+D video samples  captured with 106 distinct human subjects from 155 different camera viewpoints. 

\vspace{0.05cm}
\noindent{{\em Kinetics}}~\cite{kay2017kinetics} is a large-scale collection of 
650,000 video clips that cover 400/600/700 human action classes. 
It includes human-object interactions such as {\it playing instruments}, as well as human-human interactions such as {\it shaking hands} and {\it hugging}. As the Kinetics-400 dataset provides only the raw videos, we follow approach~\cite{stgcn2018aaai} and use the estimated joint locations in the pixel coordinate system as the input to our pipeline. To obtain the joint locations, we first resize all videos to the resolution of 340 $\times$ 256, and convert the frame rate to 30 FPS. Then we use the publicly available {\it OpenPose}~\cite{Cao_2017_CVPR} toolbox to estimate the location of 18 joints on every frame of the clips. As OpenPose  produces the 2D body joint coordinates 
and Kinetics-400 does not offer multiview or depth data, we use a network of Martinez \etal   \cite{martinez_2d23d} pre-trained on 
Human3.6M~\cite{Catalin2014Human3}, combined with the 2D OpenPose output to  estimate 3D coordinates from 2D coordinates. The 2D OpenPose  and the latter network give us $(x,y)$ and $z$ coordinates, respectively.

\vspace{0.05cm}
\noindent{\bf Evaluation protocols.} 
For the UWA3D Multiview Activity II, we use standard multi-view classification protocol~\cite{Rahmani2016, lei_tip_2019}, but we apply it to one-shot learning as the view combinations for training and testing sets are disjoint. For NTU-120, we follow the standard one-shot protocol~\cite{Liu_2019_NTURGBD120}. Based on this protocol, we create a similar one-shot protocol for NTU-60, with  50/10 action classes used for training/testing respectively. To evaluate the effectiveness of the proposed method on viewpoint alignment, we also create two new protocols on NTU-120, for which we group the whole dataset based on (i) horizontal camera views into left, center and right views, (ii) vertical camera views into top, center and bottom views. We conduct two sets of experiments on such disjoint view-wise splits: (i) using 100 action classes for training, and testing on the same 100 action classes (ii) training on 100 action classes but testing on the rest unseen 20 classes. Section \ref{app:epr} of \textbf{Suppl. Material} details
new/additional eval. protocols on NTU-60/NTU-120. 

\vspace{0.05cm}
\noindent{\bf Stereo projections.} For simulating different camera viewpoints, we  estimate the fundamental matrix $F$ (Eq.~\ref{eq:f_matrix}), 
which
relies on camera parameters. Thus, 
we use the Camera Calibrator from MATLAB to estimate  intrinsic, extrinsic and lens distortion parameters. For a given skeleton dataset, we compute the range of spatial coordinates $x$ and $y$, respectively. We then split them into 3 equally-sized groups to form roughly left, center, right views and other 3 groups for bottom, center, top views. We choose $\sim$15 frame images from each corresponding group, upload them to the Camera Calibrator, and  export  camera parameters.  We then compute the average distance/depth and height per group to estimate the camera position. On NTU-60 and NTU-120, we simply group the whole dataset into 3 cameras, which are left, center and right views, as provided in~\cite{Liu_2019_NTURGBD120}, and then we compute the average distance per camera view based on the height and distance settings given in the table in~\cite{Liu_2019_NTURGBD120}.

\subsection{Evaluations}

\comment{
\begin{figure}[t]
\centering
\vspace{-0.1cm}
\begin{subfigure}[b]{0.495\linewidth}
\includegraphics[trim=0 0 0 0, clip=true,width=0.99\linewidth]{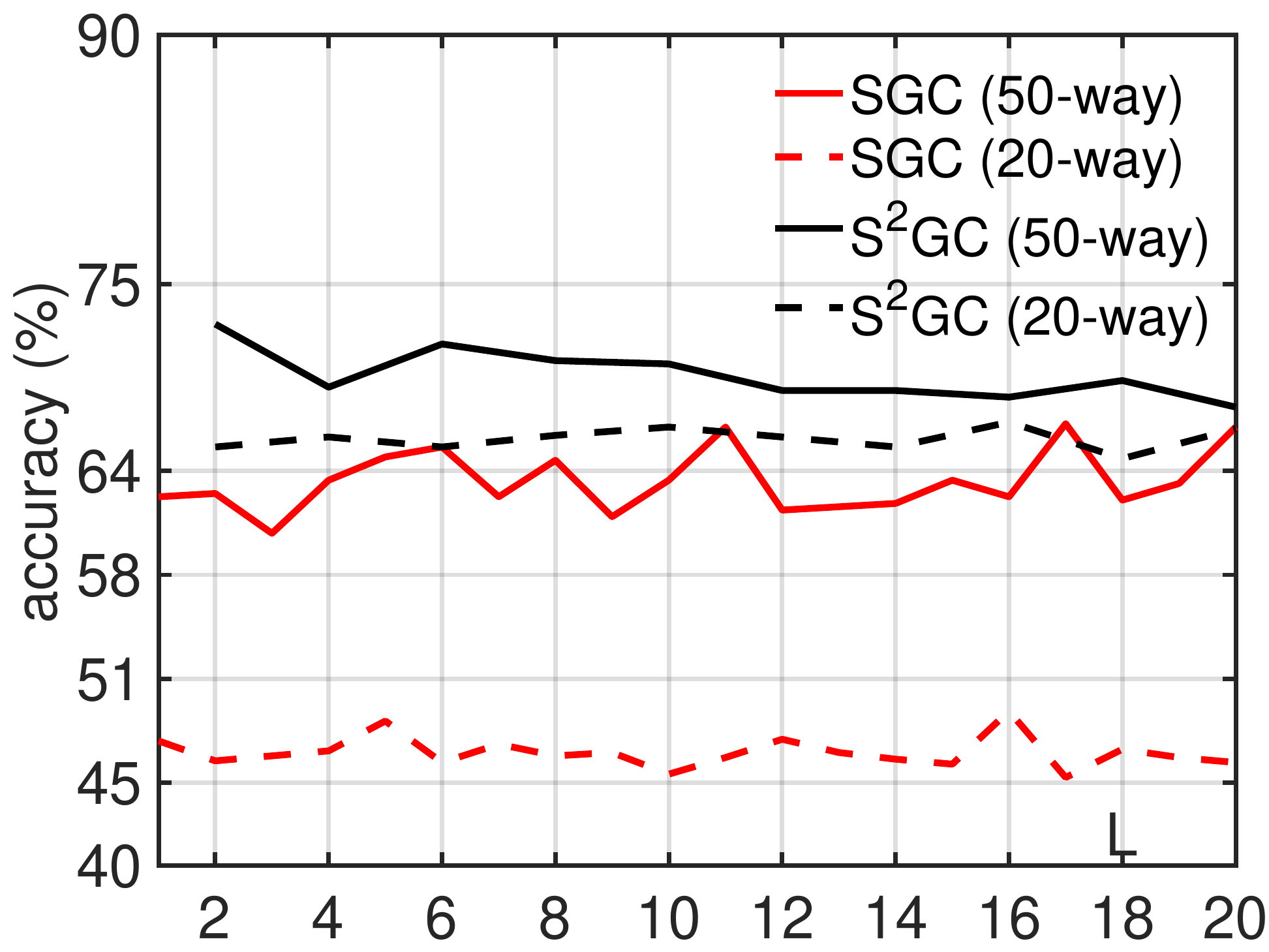}\vspace{-0.2cm}
\caption{\label{fig:degree}}
\vspace{-0.3cm}
\end{subfigure}
\begin{subfigure}[b]{0.495\linewidth}
\includegraphics[trim=0 0 0 0, clip=true,width=0.99\linewidth]{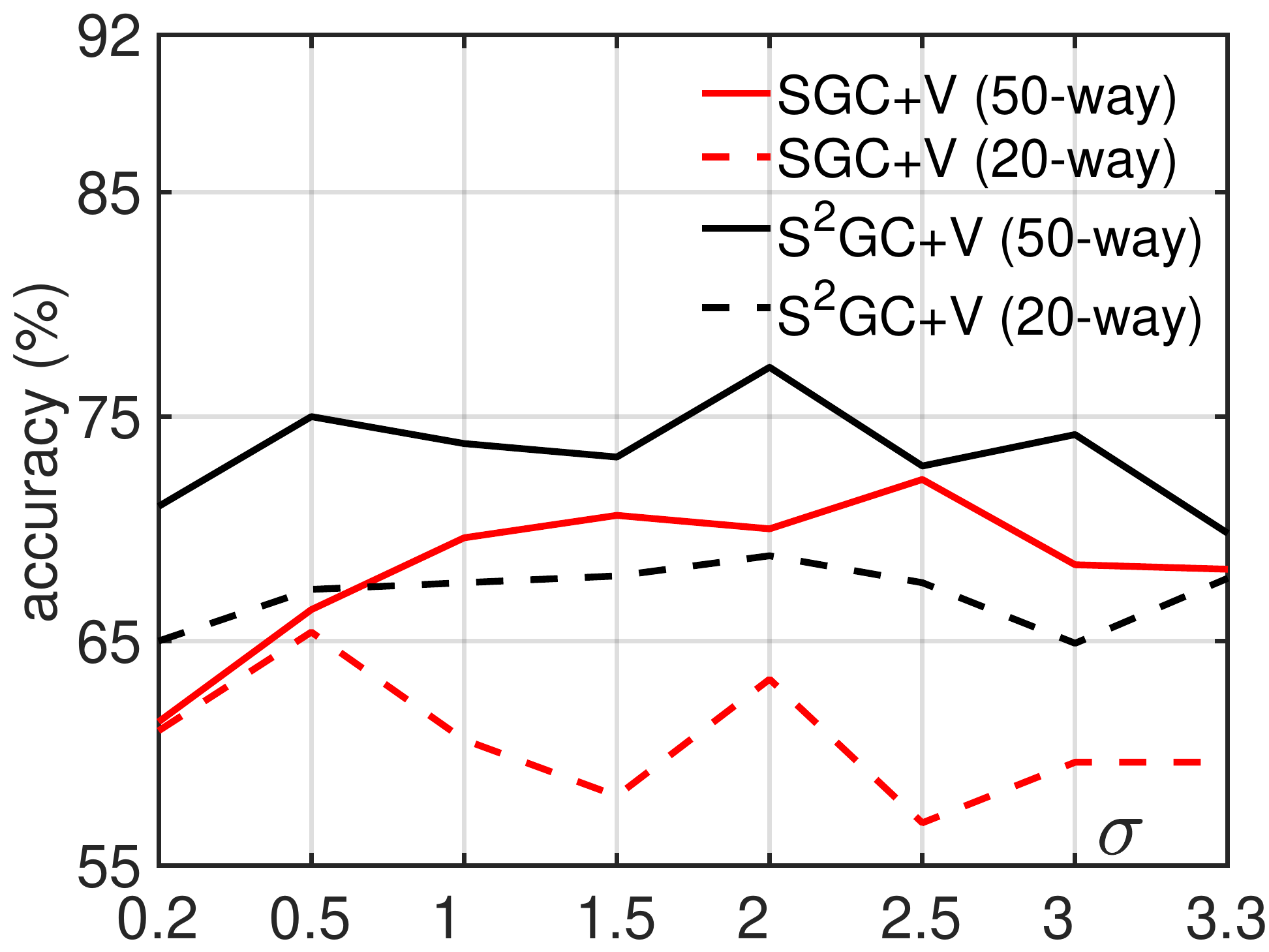}\vspace{-0.2cm}
\caption{\label{fig:sigma}}
\vspace{-0.3cm}
\end{subfigure}
\caption{
In Fig.~\ref{fig:degree}, evaluations of  $L$  for SGC and S$^2$GC. In Fig.~\ref{fig:sigma}, evaluations of  $\sigma$ of RBF distance for Eq.~\ref{eq:d_jeanie} (SGC and S$^2$GC, NTU-60).
}
\label{fig:sigma_degree}
\end{figure}
}

\begin{figure}[t]
\centering
\vspace{-0.3cm}
\begin{subfigure}[b]{0.495\linewidth}
\includegraphics[trim=0 0 0 0, clip=true,width=0.99\linewidth]{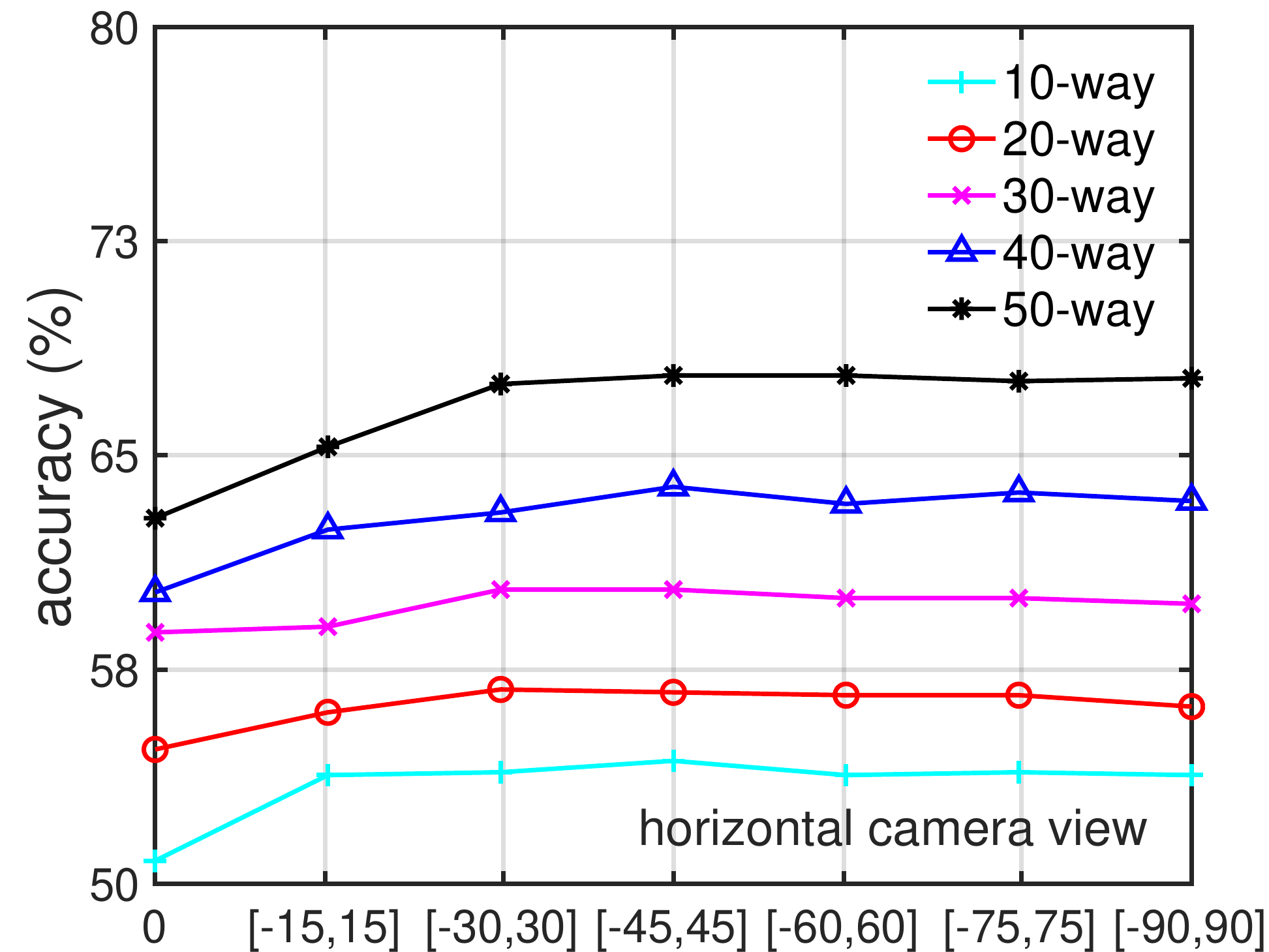}
\caption{\label{fig:sgc_h_angles}}
\vspace{-0.2cm}
\end{subfigure}
\begin{subfigure}[b]{0.495\linewidth}
\includegraphics[trim=0 0 0 0, clip=true,width=0.99\linewidth]{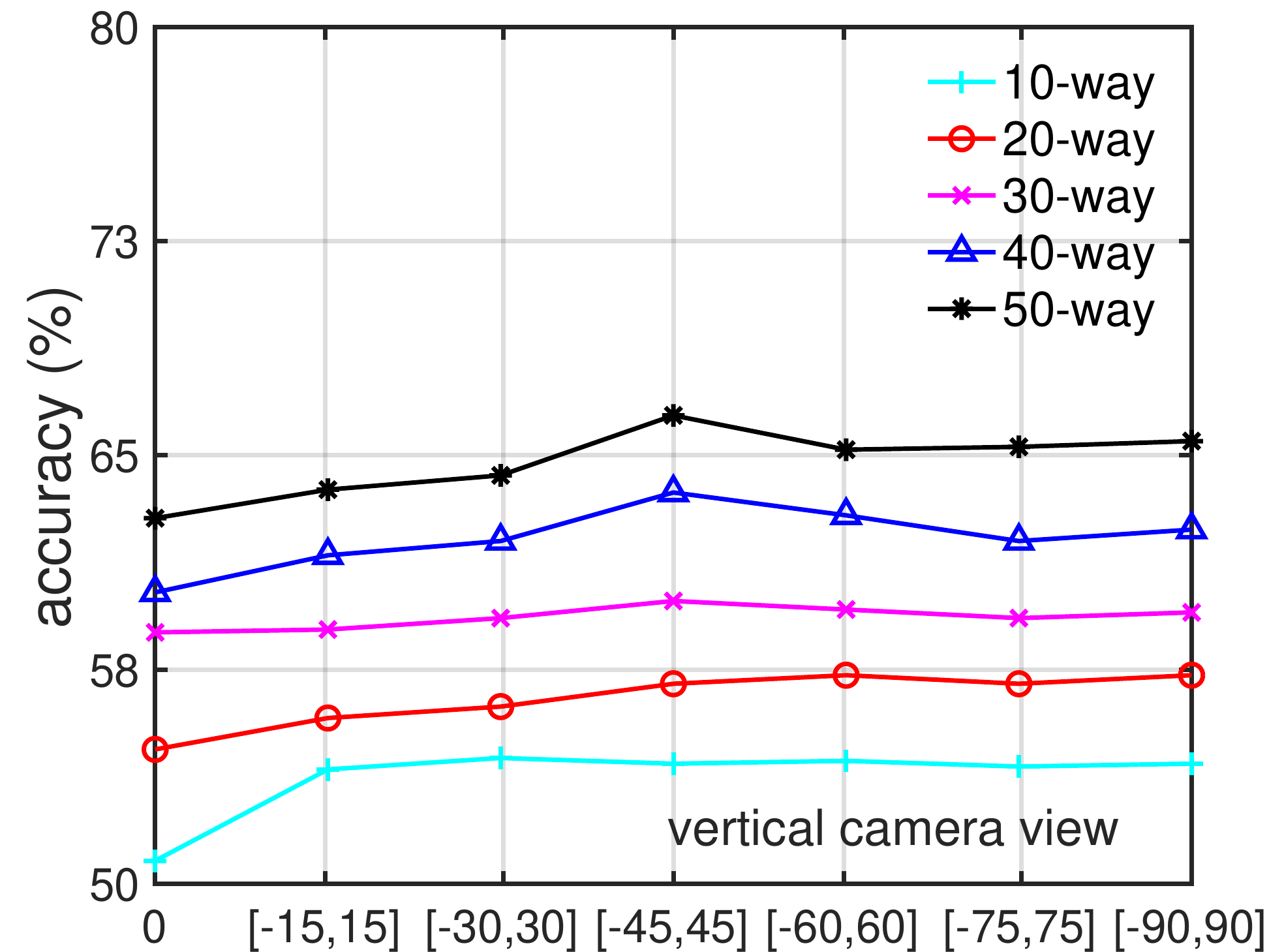}
\caption{\label{fig:sgc_v_angles}}
\vspace{-0.2cm}
\end{subfigure}
\caption{The impact of viewing angles in (Fig.~\ref{fig:sgc_h_angles}) horizontal  and (Fig.~\ref{fig:sgc_v_angles}) vertical camera views  on NTU-60.
}
\label{fig:h_v_angles}
\end{figure}

\comment{
\begin{figure}[t]
\centering
\vspace{0.1cm}
\begin{subfigure}[b]{0.495\linewidth}
\includegraphics[trim=0 0 0 0, clip=true,width=0.99\linewidth]{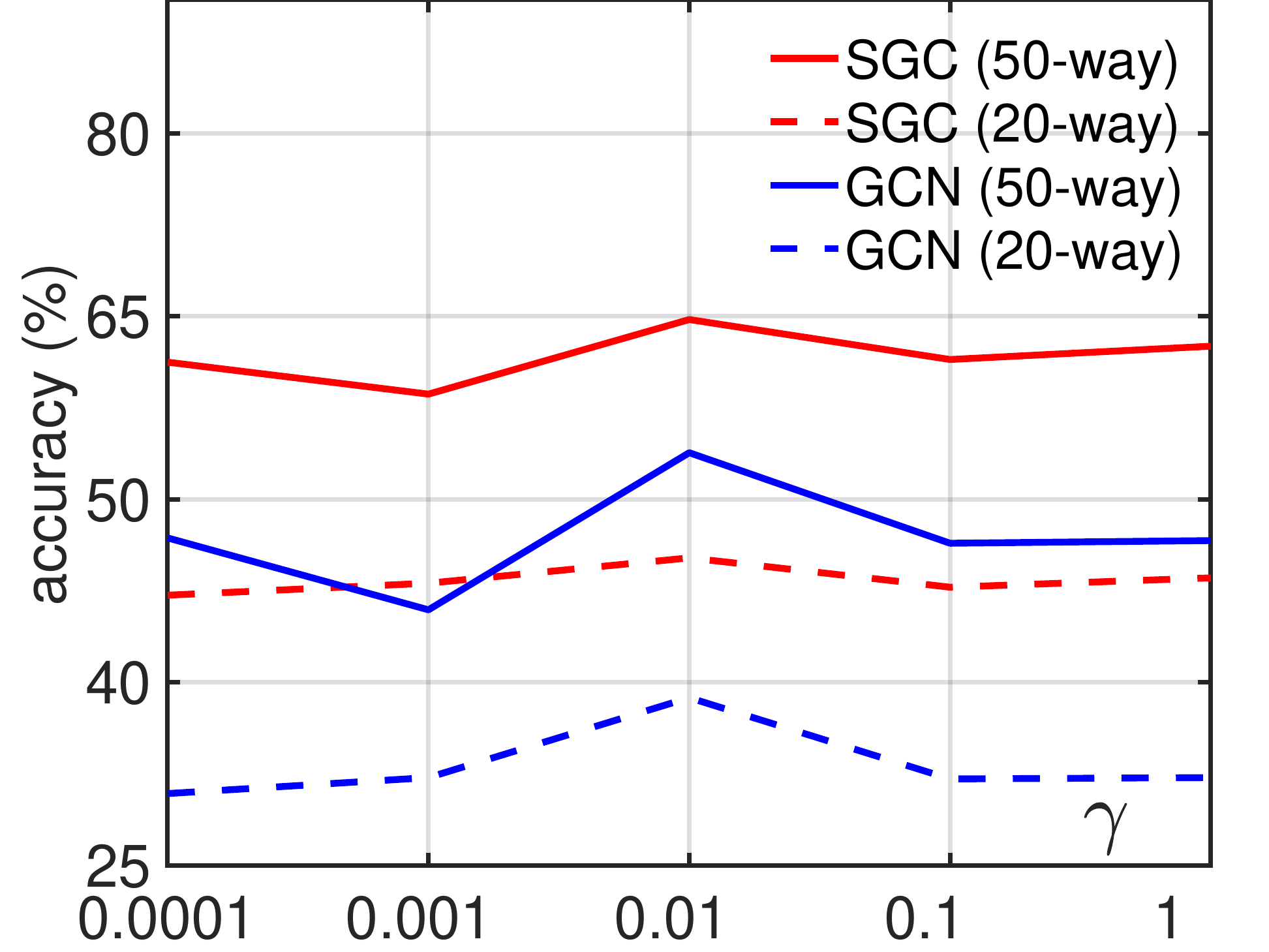}\vspace{-0.2cm}
\caption{\label{fig:sgc_gcn_gamma}}
\vspace{-0.3cm}
\end{subfigure}
\begin{subfigure}[b]{0.495\linewidth}
\includegraphics[trim=0 0 0 0, clip=true,width=0.99\linewidth]{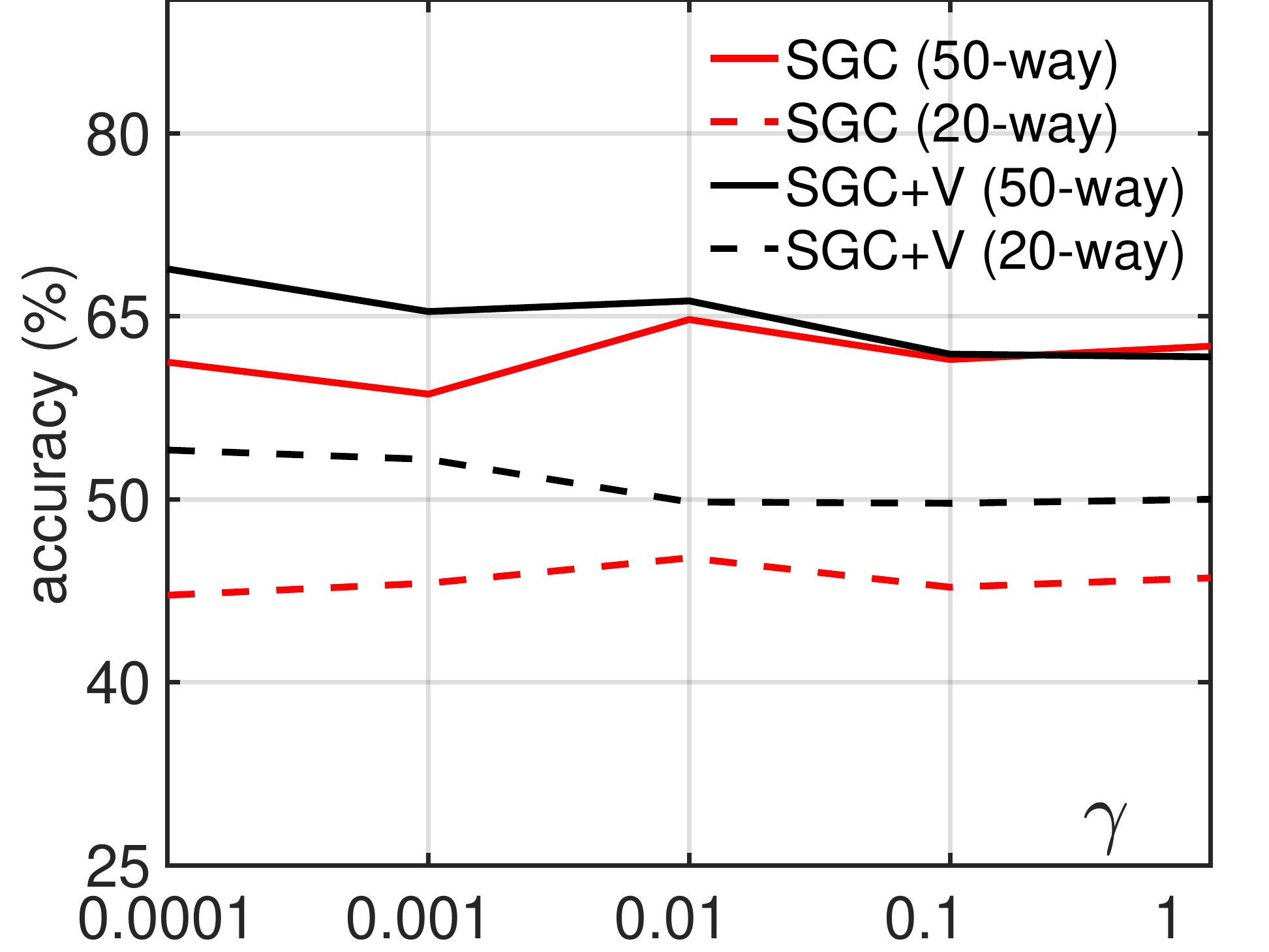}\vspace{-0.2cm}
\caption{\label{fig:sgc_vt_gamma}}
\vspace{-0.3cm}
\end{subfigure}
\caption{
In Fig.~\ref{fig:sgc_gcn_gamma}, evaluations of $\gamma$ in Eq.~\ref{eq:d_jeanie} with temporal alignment alone. In Fig.~\ref{fig:sgc_vt_gamma}, evaluations with temporal alignment alone \vs temporal-viewpoint alignment ({\em V}) on NTU-60.
}
\label{fig:gamma}
\end{figure}}

We start our experiments by investigating the GNN backbones (\textbf{Suppl. Material}, Sec.~\ref{backbone_selection} but also Table \ref{uwa3dmresults} in the main paper), their hyper-parameters (\textbf{Suppl. Material}, Sec.~\ref{block_size_strides}, \ref{eva_alpha}, \ref{eva_layer}) and camera viewpoint simulation. Below we present our final results.
\comment{
\begin{table}[tbp!]
\begin{center}
\vspace{-0.3cm}
\caption{Evaluations of backbones on 5 datasets.}
\vspace{-0.3cm}
\resizebox{\linewidth}{!}{
\renewcommand{\arraystretch}{0.9}
\setlength{\tabcolsep}{0.15em}
\begin{tabular}{c|c|c||c||c|c||c||c}
\hline
\multirow{2}{*}{} & \multicolumn{2}{c||}{MSRAction3D} & 3DAct.Pairs & \multicolumn{2}{c||}{UWA3DActivity} & NTU-60 & NTU-120\\
\cline{2-8}
 & 5-way & 10-way & 5-way & 5-way & 10-way &  50-way & 20-way \\
\hline
GCN &  56.0 $\!\pm\!$ 1.3 &  37.6 $\!\pm\!$ 1.2 &  - & 55.4 $\!\pm\!$ 0.8  & 42.4 $\!\pm\!$ 0.8 & 56.0 &  -\\
SGC & 66.0 $\!\pm\!$ 1.1 &  48.3 $\!\pm\!$ 1.1 &  69.0 $\!\pm\!$ 1.8 & 56.4 $\!\pm\!$ 0.7 & 41.6 $\!\pm\!$ 0.6 & 68.1 & 30.7  \\
APPNP & 67.2 $\!\pm\!$ 0.8 & 58.1 $\!\pm\!$ 0.8  & 69.0 $\!\pm\!$ 2.0 & 60.6 $\!\pm\!$ 1.5 & 42.4 $\!\pm\!$ 1.3 & 68.5 &  30.8 \\
S$^2$GC (Eucl.) & 68.8 $\!\pm\!$ 1.2  & 63.1 $\!\pm\!$ 0.9  & 72.2 $\!\pm\!$ 1.8 & 69.8 $\!\pm\!$ 0.7 & 58.3 $\!\pm\!$ 0.6 & 75.6 & 34.5  \\
S$^2$GC (RBF) & 73.2 $\!\pm\!$ 0.9 & 64.6 $\!\pm\!$ 0.8 &  75.6 $\!\pm\!$ 2.1 & 76.4 $\!\pm\!$ 0.7 & 58.9 $\!\pm\!$ 0.7 & 78.1 &  36.2 \\
\hline
\end{tabular}}
\label{backboneresults}
\end{center}
\vspace{-0.5cm}
\end{table}

\vspace{0.05cm}
\noindent{\bf Backbone selection}. We conduct experiments on 4 GNN backbones listed in Table~\ref{backboneresults}. S$^2$GC performs the best on all  datasets including large-scale NTU-60 and NTU-120, APPNP outperforms  SGC, and SGC outperforms GCN. We note that using the RBF-induced distance for $d_{base}(\cdot,\cdot)$ of DTW outperforms the Euclidean distance. Fig.~\ref{fig:sigma_degree} shows a comparison of using SGC and S$^2$GC on NTU-60. As shown in the figure that using S$^2$GC performs better than SGC for both 50-way and 20-way settings. We also notice that results \wrt the number of layers $L$  are more  stable for S$^2$GC than SGC. We choose $L\!=\!6$ for our experiments. Fig.~\ref{fig:sigma} shows  evaluations of $\sigma$ for the RBF-induced distance for both SGC and S$^2$GC. As $\sigma\!=\!2$ in S$^2$GC achieves the highest performance (both 50-way and 20-way), we choose S$^2$GC as the backbone  and $\sigma\!=\!2$ for the experiments.

\begin{table}[tbp!]
\vspace{-0.3cm}
\caption{
Evaluations of frame counts per temporal block $M$ and stride step $S$  on  NTU-60. $S\!=\!pM$, where $1\!-\!p$ describes the temporal block overlap percentage. Higher $p$ means fewer overlap frames between  temporal blocks.
}
\vspace{-0.25cm}
\hspace{-0.4cm}
\setlength{\tabcolsep}{2pt}
\resizebox{0.5\textwidth}{!}{
\renewcommand{\arraystretch}{0.60}
\begin{tabular}{ l | c  c  | c  c  | c  c | c  c | c  c}
\hline
 & \multicolumn{2}{c|}{$S = M$} & \multicolumn{2}{c|}{$S = 0.8M$} & \multicolumn{2}{c|}{$S = 0.6M$} & \multicolumn{2}{c|}{$S = 0.4M$} & \multicolumn{2}{c}{$S = 0.2M$} \\
\cline{2-11}
 $M$ & 50-way & 20-way & 50-way & 20-way & 50-way & 20-way & 50-way & 20-way & 50-way & 20-way \\
\hline
\hline
5 & 69.0 & 55.7 & 71.8 & 57.2 & 69.2 & 59.6 & 73.0 & 60.8 & 71.2 & 61.2\\
6 & 69.4 & 54.0 & 65.4 & 54.1 & 67.8 & 58.0 & 72.0 & 57.8 & {\bf 73.0} & {\bf 63.0} \\
8 & 67.0 & 52.7 & 67.0 & 52.5 & {\bf 73.8}& {\bf 61.8} & 67.8 & 60.3 & 68.4 & 59.4 \\
10 & 62.2 & 44.5 & 63.6 & 50.9 & 65.2 & 48.4 & 62.4 & 57.0 & 70.4 & 56.7\\
15 & 62.0 & 43.5 & 62.6 & 48.9 & 64.7 & 47.9 & 62.4 & 57.2 & 68.3 & 56.7\\
30 & 55.6 & 42.8 & 57.2 & 44.8 & 59.2 & 43.9 & 58.8 & 55.3 & 60.2 & 53.8\\
45 & 50.0 & 39.8 & 50.5 & 40.6 & 52.3 & 39.9 & 53.0 & 42.1 & 54.0 & 45.2\\
\hline
\end{tabular}}
\label{blockframe_overlap}
\vspace{-0.2cm}
\end{table}

\vspace{0.05cm}
\noindent{\bf Block size and strides.} Table~\ref{blockframe_overlap} shows  evaluations of block size $M$ and stride $S$, and indicates 
that the best performance (both 50-way and 20-way) is achieved  for smaller block size (frame count in the block) and smaller stride. 
Longer temporal blocks decrease the performance due to the temporal information not reaching the temporal alignment step. Our block encoder encodes each temporal block for learning the local temporal motions, and aggregate these block features finally to form the global temporal motion cues. Smaller stride helps capture more local motion patterns. 
Considering the computational cost and the performance, we choose $M\!=\!8$ and $S\!=\!0.6M$ for the later experiments.
}

\begin{table}[tbp!]
\vspace{-0.3cm}
\caption{Experimental results on NTU-60 (left) and NTU-120 (right) for different camera viewpoint simulations.}
\vspace{-0.6cm}
\begin{center}
\resizebox{\linewidth}{!}{\begin{tabular}{ l | c | c | c | c | c || c | c | c | c | c}
\hline
& \multicolumn{5}{c||}{NTU-60} & \multicolumn{5}{c}{NTU-120}\\
\hline
\# Training Classes & 10 & 20 & 30 & 40 & 50 & 20 & 40 & 60 & 80 & 100\\ 
\hline
\!\!\!Euler simple ($K\!+\!K'\!$)\!\!\!&  54.3 & 56.2 & 60.4 &  64.0 & 68.1 &  30.7 & 36.8 & 39.5 &  44.3 & 46.9\\ 
\!\!\!Euler ($K\!\times\!K'\!$)\!\!\!&  {\bf 60.8} & 67.4 & 67.5 &  70.3 & {\bf 75.0} &  32.9 & 39.2 & 43.5 & 48.4 & 50.2 \\  
\!\!\!CamVPC ($K\!\times\!K'\!$)\!\!\!& 59.7 & {\bf 68.7} & {\bf 68.4} & {\bf 70.4} & 73.2 & {\bf 33.1} & {\bf 40.8} & {\bf 43.7} & {\bf 48.4} & {\bf 51.4}\\
\hline

\end{tabular}}
\label{ntu60_euler_camvpc}
\end{center}
\vspace{-0.5cm}
\end{table}

\comment{
\begin{table}[tbp!]
\vspace{-0.3cm}
\caption{Experimental results on NTU-60 (top) and NTU-120 (bottom) for different camera viewpoint simulations.}
\vspace{-0.6cm}
\begin{center}
\resizebox{0.7\linewidth}{!}{\begin{tabular}{ l | c | c | c | c | c }
\hline
\# Training Classes & 10 & 20 & 30 & 40 & 50\\ 
\hline
Euler simple ($K\!+\!K'\!$) &  54.3 & 56.2 & 60.4 &  64.0 & 68.1\\ 
Euler ($K\!\times\!K'\!$) &  {\bf 60.8} & 67.4 & 67.5 &  70.3 & {\bf 75.0}\\  
CamVPC ($K\!\times\!K'\!$) & 59.7 & {\bf 68.7} & {\bf 68.4} & {\bf 70.4} & 73.2\\
\hline
\hline
\# Training Classes & 20 & 40 & 60 & 80 & 100\\ 
\hline
Euler simple ($K\!+\!K'\!$) &  30.7 & 36.8 & 39.5 &  44.3 & 46.9\\ 
Euler ($K\!\times\!K'\!$) &  32.9 & 39.2 & 43.5 & 48.4 & 50.2\\  
CamVPC ($K\!\times\!K'\!$) & {\bf 33.1} & {\bf 40.8} & {\bf 43.7} & {\bf 48.4} & {\bf 51.4}\\
\hline
\end{tabular}}
\label{ntu60_euler_camvpc}
\end{center}
\vspace{-0.5cm}
\end{table}}

\vspace{0.05cm}
\noindent{\bf Camera viewpoint simulations}. We choose 15 degrees as the step size for the viewpoints simulation. The ranges of camera azimuth/altitude are in [$-90^0$, $90^0$]. Where stated, we perform a grid search on camera azimuth/altitude with Hyperopt. 
Below, we explore the choice of the angle ranges for both horizontal and vertical views.  Fig.~\ref{fig:sgc_h_angles} and~\ref{fig:sgc_v_angles} (evaluations on the NTU-60 dataset) show that the angle range $[-45^\circ, 45^\circ]$ performs the best, and widening the range in both views does not increase the performance any further. 
Table~\ref{ntu60_euler_camvpc} shows results for the chosen range $[-45^\circ,45^\circ]$ of camera viewpoint simulations.  
Euler simple ($K\!+\!K'$) denotes a simple concatenation of features from both horizontal and vertical views, whereas Euler/CamVPC($K\!\times\!K'$) represents the grid search of all possible views.
It shows that Euler angles for the viewpoint augmentation outperform Euler simple, and CamVPC (viewpoints of query sequences are generated by the stereo projection geometry)  outperforms Euler angles in almost all the experiments on NTU-60 and NTU-120. This proves the effectiveness of using the stereo projection geometry for the viewpoint augmentation. 
More baseline experiments with/without viewpoint alignment are in Sec. \ref{sup:vall} of \textbf{Suppl. Material}. 
%

\begin{table}[tbp!]
\vspace{-0.3cm}
\caption{Results on NTU-60 (all use S$^2$GC). $\iota$-max shift is the max. viewpoint shift from block to block in JEANIE. All methods enjoy temporal alignment by soft-DTW or JEANIE (joint temporal and viewpoint alignment) except where indicated otherwise. $\psi^+ \!=\! \mu(\vd^{+})$, $\psi^- \!=\! \mu(\vd^{-})$ and $c$ is a small constant we tuned.}
\vspace{-0.6cm}
\begin{center}
\resizebox{\linewidth}{!}{\begin{tabular}{ l | c | c | c | c | c }
\hline
\# Training Classes & 10 & 20 & 30 & 40 & 50\\ 
\hline
\hline

Each frame to frontal view & 52.9 & 53.3 & 54.6 & 54.2 & 58.3\\
Each block  to frontal view & 53.9 & 56.1 & 60.1 & 63.8 & 68.0\\


Traj. aligned baseline (video-level) & 36.1 & 40.3 & 44.5 & 48.0 & 50.2\\
Traj. aligned baseline (block-level) & 52.9 & 55.8 & 59.4 & 63.6 & 66.7\\
\hline



Matching Nets~\cite{f4Matching} & 46.1 & 48.6 & 53.3 & 56.2 & 58.8\\
Matching Nets~\cite{f4Matching}+2V & 47.2 & 50.7 & 55.4 & 57.7 & 60.2\\
Prototypical Net~\cite{f1} & 47.2 & 51.1 & 54.3 & 58.9 & 63.0\\
Prototypical Net~\cite{f1}+2V & 49.8 & 53.1 & 56.7 & 60.9 & 64.3 \\
\hline
S$^2$GC (baseline, no soft-DTW) & 50.8 & 54.7 & 58.8 & 60.2 & 62.8\\ 
soft-DTW (baseline) & 53.7 & 56.2 & 60.0 & 63.9 & 67.8 \\ 
JEANIE+V(Euler)& 54.0 & 56.0 & 60.2 & 63.8 & 67.8\\
JEANIE+2V(Euler simple)&  54.3 & 56.2 & 60.4 & 64.0 & 68.1\\ 
JEANIE+2V(Euler)& 60.8 & 67.4 & 67.5 & 70.3 & 75.0\\ 
JEANIE+2V(CamVPC)& 59.7 & 68.7 & 68.4  & 70.4 & 73.2\\ 
JEANIE+2V(CamVPC+crossval.)& 63.4 & 72.4 & 73.5  & 73.2 & 78.1\\

$\qquad$ditto but loss {\em v.1}: $(\psi^+)^2 + (\psi^- \!-c)^2$ & 62.4 & 69.3 & 72.5 & 72.7 & 75.3\\
$\qquad$ditto but loss {\em v.2}: $\lvert \psi^+ \rvert_1 + \lvert \psi^- \!- c\rvert_1$ & 60.9 & 68.4 & 70.7 & 71.1 & 72.5\\
JEANIE(1-max shift)+2V(CamVPC+crossval.*)$\!\!$ & 60.8 & 70.7 & 72.5 & 72.9 & 75.2\\
JEANIE(2-max shift)+2V(CamVPC+crossval.*)$\!\!$ & {\bf 63.8} & {\bf 72.9} & {\bf 74.0} & {\bf 73.4} & {\bf 78.1}\\ 

JEANIE(3-max shift)+2V(CamVPC+crossval.*)$\!\!$ & 55.2 & 58.9 & 65.7 & 67.1 & 72.5\\
JEANIE(4-max shift)+2V(CamVPC+crossval.*)$\!\!$ & 54.5 & 57.8 & 63.5 & 65.2 & 70.4\\
\hline
2V+Transformer (baseline, no soft-DTW) & 56.0 & 64.2 & 67.3 & 70.2 & 72.9\\
2V+Transformer  & 57.3 & 66.1 & 68.8 & 72.3 & 74.0\\
JEANIE (best variant)+2V+Transformer & {\bf 65.0} & {\bf 75.2} & {\bf 76.7} & {\bf 78.9} & {\bf 80.0}\\
\hline
\end{tabular}}
\label{ntu60results}
\end{center}
\vspace{-0.5cm}
\end{table}

\vspace{0.5cm}
\noindent{\bf One-shot AR (NTU-60)}. Table~\ref{ntu60results} shows 
that 
using the viewpoint alignment simultaneously in two dimensions,  $x$ and $y$ for Euler angles, or azimuth and altitude the stereo projection geometry ({\em CamVPC}), improves the performance by 5-8\% compared to ({\em Euler simple}), a variant where the best viewpoint alignment path was chosen from the best alignment path along $x$ and the best alignment path along $y$. Euler simple is  better than  Euler with $y$ rotations only (({\em V}) includes rotations along $y$ while ({\em 2V}) includes rotations along two axes). 
When we use  HyperOpt  \cite{bergstra2015hyperopt} to search for the best angle range in which we perform the viewpoint alignment ({\em CamVPC+crossval.}), the results improve further. Enabling the viewpoint alignment for support sequences ({\em CamVPC+crossval.$^*$}) yields extra improvement. With the transformer, our best variant of JEANIE boosts the performance by $\sim$ 2\%.

We also show that aligning query and support trajectories by the angle of torso 3D joint, denoted  ({\em Traj. aligned baseline}) are not very powerful, as alluded to in Figure \ref{fig:ttrc}. We note that aligning piece-wise parts (blocks) is better than trying to align entire trajectories. In fact, aligning individual frames by torso to the frontal view ({\em Each frame to frontal view}) and aligning block average of torso direction to the frontal view ({\em Each block  to frontal view})) were marginally better. We note these baselines use soft-DTW. We provide more details of the setting and show more comparisons in Sec. \ref{sup:morebase} of \textbf{Suppl. Material}.


\begin{table}[tbp!]
\vspace{-0.3cm}
\caption{Experimental results on NTU-120 (S$^2$GC backbone). All methods enjoy temporal
alignment by soft-DTW or JEANIE (joint temporal and viewpoint
alignment) except VA \cite{Zhang_2017_ICCV,8630687} and other cited works. For VA$^*$, we used soft-DTW on temporal blocks while VA generated temporal blocks.}
\vspace{-0.6cm}
\begin{center}
\resizebox{\linewidth}{!}{\begin{tabular}{ l | c | c | c | c | c }
\hline
\# Training Classes & 20  & 40  & 60  & 80  & 100 \\ 
\hline
\hline
APSR~\cite{Liu_2019_NTURGBD120} & 29.1 & 34.8& 39.2 & 42.8 & 45.3 \\
SL-DML~\cite{2021dml}& 36.7 & 42.4 & 49.0 & 46.4& 50.9 \\
Skeleton-DML~\cite{memmesheimer2021skeletondml} & 28.6 & 37.5 & 48.6 & 48.0 & 54.2 \\
\hline

Prototypical Net+VA-RNN(aug.)~\cite{Zhang_2017_ICCV} & 25.3 & 28.6 & 32.5 & 35.2 & 38.0 \\
Prototypical Net+VA-CNN(aug.)~\cite{8630687}& 29.7 & 33.0 & 39.3 & 41.5 & 42.8 \\
Prototypical Net+VA-fusion(aug.)~\cite{8630687} & 29.8 & 33.2 & 39.5 & 41.7 & 43.0 \\
Prototypical Net+VA$^*$-fusion(aug.)~\cite{8630687} & 33.3 & 38.7 & 45.2 & 46.3 & 49.8 \\

\hline
S$^2$GC(baseline, no soft-DTW)& 30.0 & 35.9 & 39.2 & 43.6 & 46.4 \\ 
soft-DTW(baseline)& 30.3 & 37.2 & 39.7 & 44.0 & 46.8 \\ 
JEANIE+V(Euler)& 30.6 & 36.7 & 39.2 & 44.0 & 47.0\\ 
JEANIE+2V(Euler simple)& 30.7  & 36.8 & 39.5 & 44.3 &  46.9\\ 
JEANIE+2V(Euler)& 32.9 & 39.2 & 43.5 & 48.4 & 50.2 \\ 
JEANIE+2V(CamVPC)& 33.1 & 40.8 & 43.7 & 48.4 & 51.4 \\ 
JEANIE+2V(CamVPC+crossval.)& 35.8 & 41.3 & 46.3 & 49.3 & 53.9 \\ 
JEANIE+2V(CamVPC+crossval.*)& {\bf 37.2} & {\bf 43.0} & {\bf 49.2} & {\bf 50.0} & {\bf 55.2}\\
\hline
2V+Transformer(baseline, no soft-DTW) & 31.2 & 37.5 & 42.3 & 47.0 & 50.1\\
2V+Transformer & 31.6 & 38.0 & 43.2 & 47.8 & 51.3 \\
JEANIE (best variant)+2V+Transformer & {\bf 38.5} & {\bf 44.1} & {\bf 50.3} & {\bf 51.2} & {\bf 57.0}\\
\hline
\end{tabular}}
\label{ntu120results}
\end{center}
\end{table}

\vspace{0.05cm}
\noindent{\bf One-shot AR (NTU-120)}.  Table~\ref{ntu120results} shows that our proposed method achieves the best results on NTU-120, and outperforms the recent SL-DML and Skeleton-DML by 6.1\% and 2.8\% respectively (100 training classes). Note that Skeleton-DML requires the pre-trained model for the weights initialization whereas our proposed model with JEANIE is fully differentiable. 
%
%
For comparisons, we extended the view adaptive neural networks~\cite{8630687} by combining them with Prototypical Net~\cite{f1}.
VA-RNN+VA-CNN~\cite{8630687} uses 0.47M+24M parameters with random rotation augmentations while JEANIE uses 0.25--0.5M params. Their {\em rotation}+{\em translation} keys are not proven to perform smooth/optimal alignment as JEANIE. In contrast, $d_\text{JEANIE}$ performs jointly a smooth viewpoint-temporal alignment via a principled transportation plan ($\geq$3 dim. space) by design. Their use Euler angles which are a worse option (previous tables) than the camera projection of JEANIE.

We notice that ProtoNet+VA backbones is 12\% worse than our JEANIE. Even if we split skeletons into blocks to let soft-DTW perform temporal alignment of prototypes \& query, 
JEANIE is still 4--6\% better.

\begin{table}[tbp!]
\vspace{-0.4cm}
	\centering
	\caption{Experiments on 2D and 3D Kinetics-skeleton. Note that we have no results on JEANIE or Free Viewpoint Matching (FVM, see \textbf{Suppl. Material} Sec.~\ref{fvm} \&~\ref{jeanie_fvm_eval})  for 2D coordinates as these require very different viewpoint modeling than 3D coordinates.}
	\label{kinetics_results}  
	\vspace{-0.2cm}
	\resizebox{\linewidth}{!}{\begin{tabular}{c|c|c|c|c|c}  
		\hline  
		&S$^2$GC(baseline,& soft-DTW & {\it FVM} & {\it JEANIE} & JEANIE (best variant)\\
		&no soft-DTW) & (baseline) & + 2V & + 2V & +2V+Transformer\\%
		\hline
		\hline
		2D skel. & 32.8 & 34.7 & - & - & -\\
		3D skel. & 35.9 & 39.6 & 44.1 & {\bf 50.3} & {\bf 52.5}\\
		\hline
	\end{tabular}}
	\vspace{-0.3cm}
\end{table}

\vspace{0.05cm}
\noindent{\bf JEANIE on the Kinetics-skeleton}. We evaluate our proposed model on both 2D and 3D Kinetics-skeleton. We split the whole dataset into 200 actions for training, and the rest half for testing. As we are unable to estimate the camera location, we simply use Euler angles for the camera viewpoint simulation. 
Table~\ref{kinetics_results} shows that using 3D skeletons outperforms the use of 2D skeletons by 3-4\%, and  JEANIE outperforms the baseline (temporal alignment only) and Free Viewpoint Matching (FVM, for every step of DTW, seeks the best local viewpoint alignment, thus realizing non-smooth temporal-viewpoint path in contrast to JEANIE) by around 5\% and 6\%,  respectively. With the transformer, JEANIE  further boosts results by  2\%.

\begin{table}[tbp!]
\begin{center}
\vspace{-0.3cm}
\caption{Experiments  on the UWA3D Multiview Activity II.
}
\vspace{-0.3cm}
\resizebox{\linewidth}{!}{\begin{tabular}{ l|  c  c  c  c  c c  c  c  c  c c  c  |c  }
\hline
\!\!\!Training view\!\!\! & \multicolumn{2}{c}{$V_1$ \& $V_2$} & \multicolumn{2}{c}{$V_1$ \& $V_3$} & \multicolumn{2}{c}{$V_1$ \& $V_4$} & \multicolumn{2}{c}{$V_2$ \& $V_3$} & \multicolumn{2}{c}{$V_2$ \& $V_4$} & \multicolumn{2}{c|}{$V_3$ \& $V_4$} & Mean\\
\cline{1-13}
\!\!\!Testing view & $V_3$ & $V_4$ & $V_2$ & $V_4$ & $V_2$ & $V_3$ & $V_1$ & $V_4$ & $V_1$ & $V_3$ & $V_1$ & $V_2$ & {}\\
\hline
\hline
GCN &36.4 & 26.2 &20.6 & 30.2 & 33.7 & 22.4 & 43.1 & 26.6 & 16.9 & 12.8 & 26.3 & 36.5 & 27.6 \\
\hline
SGC&40.9&60.3&44.1&52.6&48.5&38.7&50.6&52.8&52.8&37.2&57.8&49.6&48.8\\
\!\!\!+soft-DTW\!\!\!&43.9&60.8&48.1&54.6&52.6&45.7&54.0&58.2&56.7&40.2&60.2&51.1&52.2\\
\!\!\!+JEANIE+2V\!\!\!&47.0&62.8&50.4&57.8&53.6&47.0&57.9&62.3&57.0&44.8&61.7&52.3&54.6\\
\hline
APPNP&42.9&61.9&47.8&58.7&53.8&44.0&52.3&60.3&55.1&38.2&58.3&47.9&51.8\\
\!\!\!+soft-DTW\!\!\!&44.3&63.2&50.7&62.3&53.9&45.0&56.9&62.8&56.4&39.3&60.1&51.9&53.9\\
\!\!\!+JEANIE+2V\!\!\!&46.8&64.6&51.3&65.1&54.7&46.4&58.2&65.1&58.8&43.9&60.3&52.5&55.6\\
\hline
S$^2$GC&45.5&64.4&46.8&61.6&49.5&43.2&57.3&61.2&51.0&42.9&57.0&49.2&52.5\\
\!\!\!+soft-DTW\!\!\!&48.2&67.2&51.2&67.0&53.2&46.8&62.4&66.2&57.8&45.0&62.2&53.0&56.7\\
\!\!\!+JEANIE+2V\!\!\!&{\bf 55.3}&{\bf 70.2}&{\bf 61.4}&{\bf 72.5}&{\bf 60.9}&{\bf 50.8}&{\bf 66.4}&{\bf 73.9}&{\bf 68.8}&{\bf 57.2}&{\bf 66.7}&{\bf 60.2}&{\bf 63.7}\\
\hline
\end{tabular}}
\label{uwa3dmresults}
\end{center}
\vspace{-0.2cm}
\end{table}

\begin{table}
\vspace{-0.3cm}
\caption{Results on NTU-120 (multiview classification).
}
\vspace{-0.6cm}
\begin{center}
\resizebox{\linewidth}{!}{\begin{tabular}{ l | c | c | c || c | c  | c }
\hline
Training view & bott. & bott. & bott.\& cent. & left & left & left \& cent.\\
\hline
Testing view & cent. & top & top & cent. & right & right \\
\hline
100/same 100& 81.5 & 79.2 & 83.9 & 67.7 & 66.9 & 79.2\\
100/novel 20 & 67.8 & 65.8 & 70.8 & 59.5 & 55.0 & 62.7 \\
\hline
\end{tabular}}
\label{ntu120results_view}
\end{center}
\vspace{-0.5cm}
\end{table}

\vspace{0.05cm}
\noindent{\bf Few-shot multiview classification}. Table~\ref{uwa3dmresults} (UWA3D Multiview Activity II) shows that adding temporal alignment to SGC, APPNP and S$^2$GC improves the performance, and the big performance gain is obtained via further adding the viewpoint alignment. As this dataset is challenging in recognizing the actions from a novel view point, our proposed method performs consistently well on all different combinations of training/testing viewpoint variants. This is predictable as our method align both temporal and camera viewpoint which allows a robust classification. 

Table~\ref{ntu120results_view} shows the experimental results on the NTU-120. We notice that adding more camera viewpoints to the training process helps the multiview classification, \eg using bottom and center views for training and top view for testing, and using left and center views for training and the right view for testing, and the performance gain is more than 4\% (({\em same 100}) means the same train/test classes but different views). We also notice that even though we test on 20 novel classes ({\em novel 20}) which are never used in the training set, we still achieve 62.7\% and 70.8\% for multiview classification in horizontal and vertical camera viewpoints.

\section{Conclusions}
We have proposed a Few-shot Action Recognition approach for learning on 3D skeletons via JEANIE. We have demonstrated that the joint alignment of temporal blocks and simulated viewpoints of skeletons between support-query sequences is efficient in the meta-learning setting where the alignment has to be performed on new action classes under the low number of samples (in contrast to classification pipelines such as celebrated \cite{8630687}). Our experiments have shown that using the stereo camera geometry is more efficient than simply generating multiple views by Euler angles in the meta-learning regime. Most importantly, we have contributed a novel FSAR approach that learns on articulated 3D body joints to the very limited FSAR family.

{\small
\bibliographystyle{ieee_fullname}
\bibliography{jeanie}
}

\newpage
\appendix
\title{3D Skeleton-based Few-shot Action Recognition with JEANIE is not so Na\"ive (Supplementary Material)\vspace{-0.3cm}}

\author{%
  Lei Wang$^{\dagger,\S}$, \quad Jun Liu$^{\spadesuit}$, \quad Piotr Koniusz\textsuperscript{\textasteriskcentered}$^{\!,\S,\dagger}$\\\vspace{0.3cm}
  \!\!\!\!\!\!$^{\dagger}$The Australian National University \!\!\quad \!\!
  $^{\spadesuit}$Singapore University of Technology and Design \!\!\quad\!\! $^\S$Data61/CSIRO \\
	\vspace{-0.5cm}
  \{lei.wang, piotr.koniusz\}@data61.csiro.au \quad jun\_liu@sutd.edu.sg \\
	\vspace{-0.3cm}
}

\maketitle

\renewcommand{\dblfloatpagefraction}{0.99}

\renewcommand{\topfraction}{0.99} 
\renewcommand{\bottomfraction}{0.99} 
\renewcommand{\textfraction}{0.01} 
\renewcommand{\floatpagefraction}{0.99}

\setcounter{topnumber}{4}
\setcounter{bottomnumber}{4}
\setcounter{totalnumber}{8}

\section{Datasets and their statistics}
\label{app:ds}

Table \ref{datasets} contains statistics of datasets used in our experiments. 
Smaller datasets below are used for the backbone selection and ablations: 

\vspace{0.05cm}
\begin{itemize}

\item {\em{MSRAction3D}}~\cite{Li2010} is an older AR datasets captured with the Kinect depth camera. It contains 20 human sport-related activities such as {\it jogging}, {\it golf swing} and {\it side boxing}. 

\item {{\em 3D Action Pairs}}~\cite{Oreifej2013} contains 6 selected pairs of actions that have very similar motion trajectories \eg, {\it put on a hat} and {\it take off a hat}, {\it pick up a box} and {\it put down a box}, \etc. 

\item {{\em UWA3D Activity}}~\cite{RahmaniHOPC2014} has 30 actions performed by 10 people of various height at different speeds in cluttered scenes. 
\end{itemize}

As MSRAction3D, 3D Action Pairs, and UWA3D Activity have not been used in FSAR, we created 10 training/testing splits, by choosing half of class concepts for training, and half for testing per split per dataset. Training splits were further subdivided for crossvalidation.

Sec.~\ref{small_proto} details the class concepts per split for small datasets.

\begin{table*}[tbp!]
\vspace{-0.3cm}
\caption{Seven publicly available benchmark datasets which we use for FSAR.}
\vspace{-0.5cm}
\begin{center}
\resizebox{0.85\textwidth}{!}{\begin{tabular}{ l  c | c | c | c | c | c | c | c | c }
\hline
 Datasets & & Year & Classes & Subjects & \#views & \#clips & Sensor & Modalities & \#joints \\ 
\hline
\hline
MSRAction3D~& \cite{Li2010} & 2010 & 20 & 10 & 1 & 567 & Kinect v1 & Depth + 3DJoints & 20\\
3D Action Pairs~& \cite{Oreifej2013} & 2013 & 12 & 10 & 1 & 360 & Kinect v1 & RGB + Depth + 3DJoints & 20\\
UWA3D Activity~& \cite{RahmaniHOPC2014} & 2014 & 30 & 10 & 1 & 701 & Kinect v1 & RGB + Depth + 3DJoints & 15\\
UWA3D Multiview Activity II& \cite{Rahmani2016} & 2015 & 30 & 9 & 4 & 1,070 & Kinect v1 & RGB + Depth + 3DJoints & 15\\
NTU RGB+D~& \cite{Shahroudy_2016_NTURGBD} & 2016 & 60 & 40 & 80 & 56,880 & Kinect v2 & RGB + Depth + IR + 3DJoints & 25\\
NTU RGB+D 120~& \cite{Liu_2019_NTURGBD120} & 2019 & 120 & 106 & 155 & 114,480 & Kinect v2 & RGB + Depth + IR + 3DJoints & 25\\
Kinetics-skeleton~& \cite{stgcn2018aaai} & 2018 & 400 & - & - & $\sim$ 300,000 & - & RGB + 2DJoints & 18 \\
\hline
\end{tabular}}
\label{datasets}
\end{center}
\vspace{-0.4cm}
\end{table*}

\section{Backbone selection and hyperparameter evaluation}
\label{backbone_hyperpara}

\begin{figure}[tbp!]
\vspace{-0.3cm}
\centering
\begin{subfigure}[b]{0.495\linewidth}
\includegraphics[trim=0 0 0 0, clip=true,width=0.99\linewidth]{imgs/sgc_degree.pdf}\vspace{-0.1cm}
\caption{\label{fig:degree}}
\vspace{-0.3cm}
\end{subfigure}
\begin{subfigure}[b]{0.495\linewidth}
\includegraphics[trim=0 0 0 0, clip=true,width=0.99\linewidth]{imgs/sgc_rbf_gamma.pdf}\vspace{-0.1cm}
\caption{\label{fig:sigma}}
\vspace{-0.3cm}
\end{subfigure}
\caption{
Evaluations of $L$ and $\sigma$. Fig.~\ref{fig:degree}: $L$ for SGC and S$^2$GC. Fig.~\ref{fig:sigma}: $\sigma$ of RBF distance for Eq.~\ref{eq:d_jeanie} (SGC and S$^2$GC, NTU-60).
}
\label{fig:sigma_degree}
\end{figure}

\comment{
\begin{figure}[t]
\centering
\begin{subfigure}[b]{0.495\linewidth}
\includegraphics[trim=0 0 0 0, clip=true,width=0.99\linewidth]{imgs/sgc_h_angles.pdf}\vspace{-0.1cm}
\caption{\label{fig:sgc_h_angles}}
\vspace{-0.3cm}
\end{subfigure}
\begin{subfigure}[b]{0.495\linewidth}
\includegraphics[trim=0 0 0 0, clip=true,width=0.99\linewidth]{imgs/sgc_v_angles.pdf}\vspace{-0.1cm}
\caption{\label{fig:sgc_v_angles}}
\vspace{-0.3cm}
\end{subfigure}
\caption{The impact of viewing angles in (Fig.~\ref{fig:sgc_h_angles}) horizontal  and (Fig.~\ref{fig:sgc_v_angles}) vertical camera views  on NTU-60.
}
\label{fig:h_v_angles}
\end{figure}}

\begin{figure}[tbp!]
\centering
\vspace{-0.3cm}
\begin{subfigure}[b]{0.495\linewidth}
\includegraphics[trim=0 0 0 0, clip=true,width=0.99\linewidth]{imgs/sgc_gcn_t_gamma.pdf}\vspace{-0.1cm}
\caption{\label{fig:sgc_gcn_gamma}}
\vspace{-0.3cm}
\end{subfigure}
\begin{subfigure}[b]{0.495\linewidth}
\includegraphics[trim=0 0 0 0, clip=true,width=0.99\linewidth]{imgs/sgc_tv_gamma.pdf}\vspace{-0.1cm}
\caption{\label{fig:sgc_vt_gamma}}
\vspace{-0.3cm}
\end{subfigure}
\caption{
Evaluations \wrt $\gamma$. Fig.~\ref{fig:sgc_gcn_gamma}: $\gamma$ in Eq.~\ref{eq:d_jeanie} with the temporal alignment alone.  Fig.~\ref{fig:sgc_vt_gamma}: comparisons of temporal alignment alone \vs temporal-viewpoint alignment ({\em V}) on NTU-60.
}
\label{fig:gamma}
\end{figure}

\begin{table}[tbp!]
\begin{center}
\vspace{-0.3cm}
\caption{Evaluations of backbones on 5 datasets.}
\vspace{-0.3cm}
\resizebox{\linewidth}{!}{
\renewcommand{\arraystretch}{0.9}
\setlength{\tabcolsep}{0.15em}
\begin{tabular}{c|c|c||c||c|c||c||c}
\hline
\multirow{2}{*}{} & \multicolumn{2}{c||}{MSRAction3D} & 3DAct.Pairs & \multicolumn{2}{c||}{UWA3DActivity} & NTU-60 & NTU-120\\
\cline{2-8}
 & 5-way & 10-way & 5-way & 5-way & 10-way &  50-way & 20-way \\
\hline
GCN &  56.0 $\!\pm\!$ 1.3 &  37.6 $\!\pm\!$ 1.2 &  - & 55.4 $\!\pm\!$ 0.8  & 42.4 $\!\pm\!$ 0.8 & 56.0 &  -\\
SGC & 66.0 $\!\pm\!$ 1.1 &  48.3 $\!\pm\!$ 1.1 &  69.0 $\!\pm\!$ 1.8 & 56.4 $\!\pm\!$ 0.7 & 41.6 $\!\pm\!$ 0.6 & 68.1 & 30.7  \\
APPNP & 67.2 $\!\pm\!$ 0.8 & 58.1 $\!\pm\!$ 0.8  & 69.0 $\!\pm\!$ 2.0 & 60.6 $\!\pm\!$ 1.5 & 42.4 $\!\pm\!$ 1.3 & 68.5 &  30.8 \\
S$^2$GC (Eucl.) & 68.8 $\!\pm\!$ 1.2  & 63.1 $\!\pm\!$ 0.9  & 72.2 $\!\pm\!$ 1.8 & 69.8 $\!\pm\!$ 0.7 & 58.3 $\!\pm\!$ 0.6 & 75.6 & 34.5  \\
S$^2$GC (RBF) & 73.2 $\!\pm\!$ 0.9 & 64.6 $\!\pm\!$ 0.8 &  75.6 $\!\pm\!$ 2.1 & 76.4 $\!\pm\!$ 0.7 & 58.9 $\!\pm\!$ 0.7 & 78.1 &  36.2 \\
\hline
\end{tabular}}
\label{backboneresults}
\end{center}
\end{table}

\subsection{Backbone selection}
\label{backbone_selection}

We conduct experiments on 4 GNN backbones listed in Table~\ref{backboneresults}. S$^2$GC performs the best on all  datasets including large-scale NTU-60 and NTU-120, APPNP outperforms  SGC, and SGC outperforms GCN. We note that using the RBF-induced distance for $d_{base}(\cdot,\cdot)$ of DTW outperforms the Euclidean distance. Fig.~\ref{fig:sigma_degree} shows a comparison of using SGC and S$^2$GC on NTU-60. As shown in the figure that using S$^2$GC performs better than SGC for both 50-way and 20-way settings. We also notice that results \wrt the number of layers $L$  are more  stable for S$^2$GC than SGC. We choose $L\!=\!6$ for our experiments. Fig.~\ref{fig:sigma} shows  evaluations of $\sigma$ for the RBF-induced distance for both SGC and S$^2$GC. As $\sigma\!=\!2$ in S$^2$GC achieves the highest performance (both 50-way and 20-way), we choose S$^2$GC as the backbone  and $\sigma\!=\!2$ for the experiments.

\begin{table}[tbp!]
\vspace{-0.3cm}
\caption{
Evaluations of frame counts per temporal block $M$ and stride step $S$  on  NTU-60. $S\!=\!pM$, where $1\!-\!p$ describes the temporal block overlap percentage. Higher $p$ means fewer overlap frames between  temporal blocks.
}
\vspace{-0.25cm}
\hspace{-0.4cm}
\setlength{\tabcolsep}{2pt}
\resizebox{0.5\textwidth}{!}{
\renewcommand{\arraystretch}{0.60}
\begin{tabular}{ l | c  c  | c  c  | c  c | c  c | c  c}
\hline
 & \multicolumn{2}{c|}{$S = M$} & \multicolumn{2}{c|}{$S = 0.8M$} & \multicolumn{2}{c|}{$S = 0.6M$} & \multicolumn{2}{c|}{$S = 0.4M$} & \multicolumn{2}{c}{$S = 0.2M$} \\
\cline{2-11}
 $M$ & 50-way & 20-way & 50-way & 20-way & 50-way & 20-way & 50-way & 20-way & 50-way & 20-way \\
\hline
\hline
5 & 69.0 & 55.7 & 71.8 & 57.2 & 69.2 & 59.6 & 73.0 & 60.8 & 71.2 & 61.2\\
6 & 69.4 & 54.0 & 65.4 & 54.1 & 67.8 & 58.0 & 72.0 & 57.8 & {\bf 73.0} & {\bf 63.0} \\
8 & 67.0 & 52.7 & 67.0 & 52.5 & {\bf 73.8}& {\bf 61.8} & 67.8 & 60.3 & 68.4 & 59.4 \\
10 & 62.2 & 44.5 & 63.6 & 50.9 & 65.2 & 48.4 & 62.4 & 57.0 & 70.4 & 56.7\\
15 & 62.0 & 43.5 & 62.6 & 48.9 & 64.7 & 47.9 & 62.4 & 57.2 & 68.3 & 56.7\\
30 & 55.6 & 42.8 & 57.2 & 44.8 & 59.2 & 43.9 & 58.8 & 55.3 & 60.2 & 53.8\\
45 & 50.0 & 39.8 & 50.5 & 40.6 & 52.3 & 39.9 & 53.0 & 42.1 & 54.0 & 45.2\\
\hline
\end{tabular}}
\label{blockframe_overlap}
\vspace{-0.2cm}
\end{table}

\subsection{Block size and strides}
\label{block_size_strides}

Table~\ref{blockframe_overlap} shows  evaluations of block size $M$ and stride $S$, and indicates 
that the best performance (both 50-way and 20-way) is achieved  for smaller block size (frame count in the block) and smaller stride. 
Longer temporal blocks decrease the performance due to the temporal information not reaching the temporal alignment step. Our block encoder encodes each temporal block for learning the local temporal motions, and aggregate these block features finally to form the global temporal motion cues. Smaller stride helps capture more local motion patterns. 
Considering the computational cost and the performance, we choose $M\!=\!8$ and $S\!=\!0.6M$ for the later experiments.

Below, we evaluate two main hyperparameters in S$^2$GC, which are $\alpha$ and the number of layers $L$.

\begin{figure}[tbp!]
\centering
\begin{subfigure}[b]{0.495\linewidth}
\includegraphics[trim=0 0 0 0, clip=true,width=0.99\linewidth]{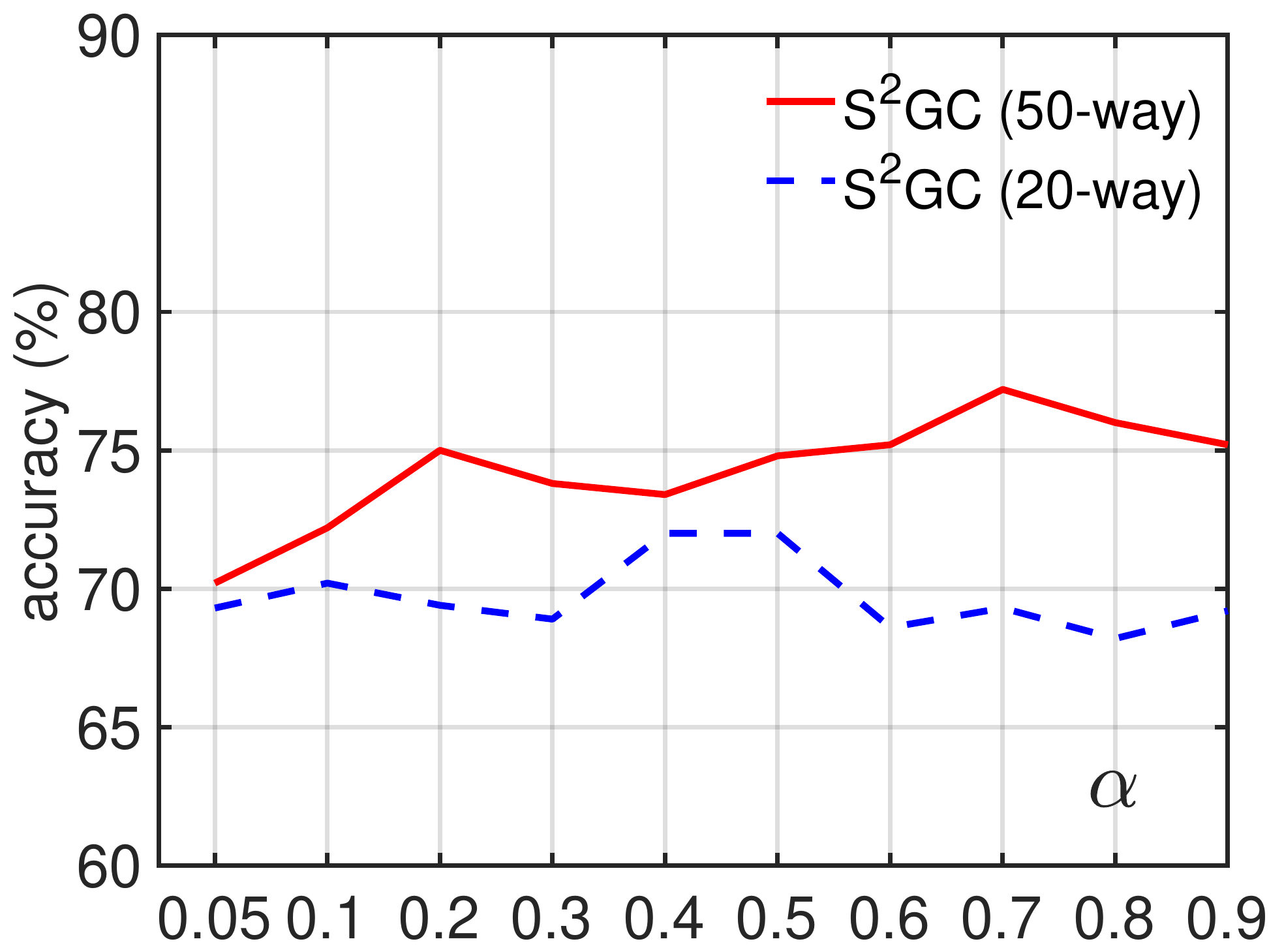}\vspace{-0.2cm}
\caption{\label{fig:ssgc_alpha}}
\vspace{-0.3cm}
\end{subfigure}
\begin{subfigure}[b]{0.495\linewidth}
\includegraphics[trim=0 0 0 0, clip=true,width=0.99\linewidth]{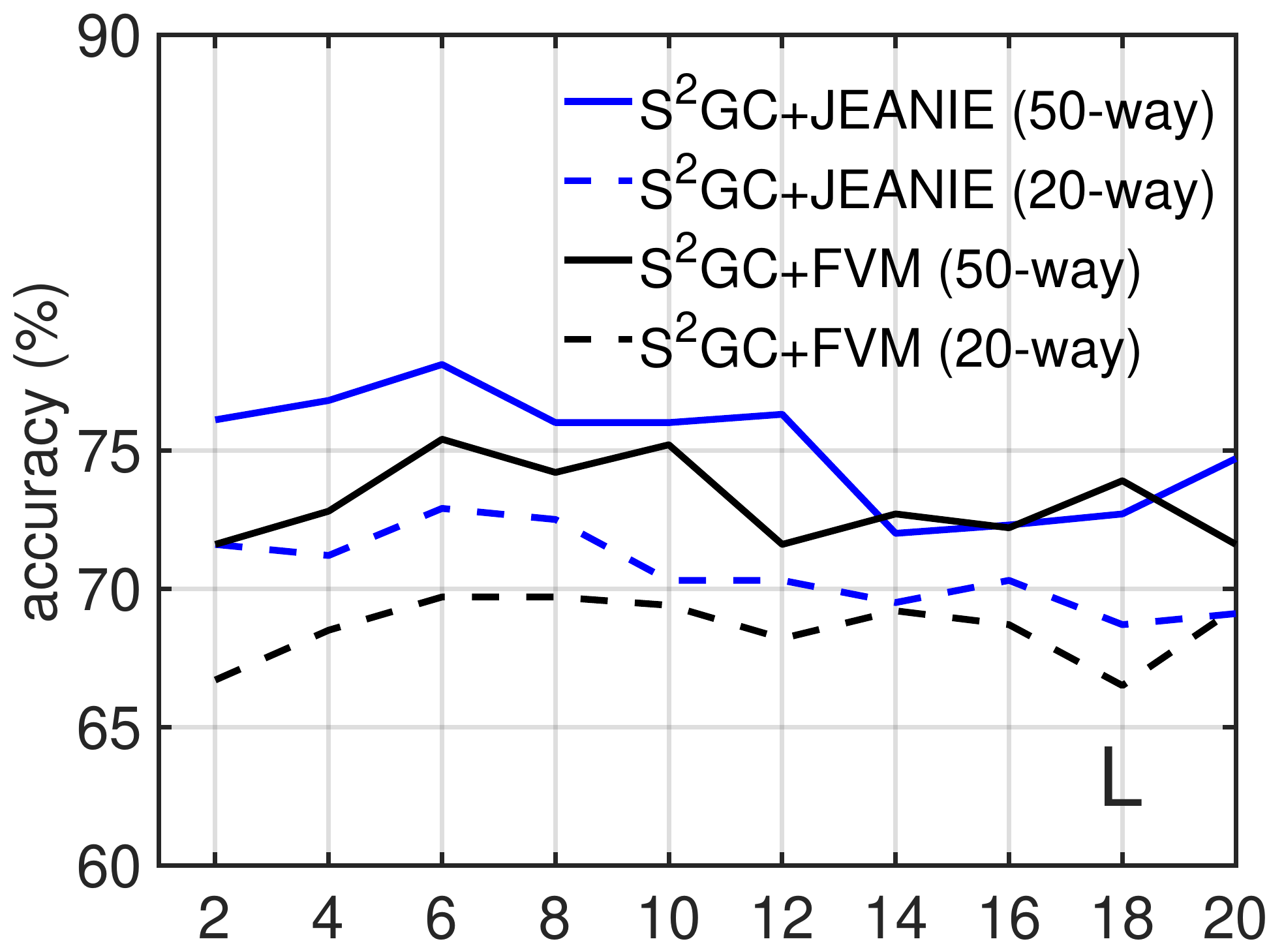}\vspace{-0.2cm}
\caption{\label{fig:ssgc_degree}}
\vspace{-0.3cm}
\end{subfigure}
\caption{Evaluations of $\alpha$ (Fig.~\ref{fig:ssgc_alpha}) and the number of layers $L$ (Fig.~\ref{fig:ssgc_degree}) for S$^2$GC on NTU-60.
}
\label{fig:ssgc_alpha_degree}
\end{figure}

\subsection{Evaluations of viewpoint alignment}
\label{sup:vall}

Fig.~\ref{fig:gamma} shows comparisons  with temporal-viewpoint alignment ({\em V}) \vs temporal alignment alone on NTU-60. 

Fig.~\ref{fig:sgc_gcn_gamma} shows  evaluations of $\gamma$ without the viewpoint alignment. Fig.~\ref{fig:sgc_vt_gamma} shows that temporal-viewpoint alignment ({\em V}) brings around 5\% (20- and 50-way protocols, $\gamma\!=\!0.0001$). 

\subsection{Evaluations \wrt ${\alpha}$}
\label{eva_alpha}
Figure~\ref{fig:ssgc_alpha} shows the evaluations of $\alpha$ for the S$^2$GC backbone. As shown in the plot, for 50-way protocol, the best performance is achieved when $\alpha\!=\!0.7$ ($\alpha\!=\!0.5$  is the second best performer for the 50-way protocol)). For the 20-way protocol, the top performer is $\alpha\!=\!0.4$ or $\alpha\!=\!0.5$. Thus, we chose $\alpha\!=\!0.5$ in our experiments. Please note we observed the same trend on the validation split.

\subsection{Evaluations \wrt the number of layers ${L}$}
\label{eva_layer}

Figure~\ref{fig:ssgc_degree} shows the performance \wrt the number of layers $L$ used by S$^2$GC and S$^2$GC+JEANIE. As shown in this plot,  when $L\!=\!6$, S$^2$GC with JEANIE performs the best for both 20- and 50-way experiments. For the Free Viewpoint Matching using S$^2$GC (S$^2$GC+FVM), the performance is not as stable as in the case of S$^2$GC+JEANIE. 


\section{More baselines on NTU-60}
\label{sup:morebase}

\begin{table}[tbp!]
\vspace{-0.3cm}
\caption{Experimental results on NTU-60.}
\vspace{-0.6cm}
\begin{center}
\resizebox{\linewidth}{!}{\begin{tabular}{ l | c | c | c | c | c }
\hline
\# Training Classes & 10 & 20 & 30 & 40 & 50\\ 
\hline
\hline


Matching Nets~\cite{f4Matching} (skeleton to image tensor)& 26.7 & 30.6 & 32.9 & 36.4 & 39.9\\
Proto. Net~\cite{f1} (skeleton to image tensor)& 30.6 & 33.9 & 36.8 & 40.2 & 43.0\\
Proto. Net (per block image tensor, temp. align.)& 40.4 & 42.4 & 45.2 & 49.0 & 50.3\\
\!\!\!\!\!Proto. Net (per block image tensor, temp. \& view. align.) & 41.6 & 43.0 & 47.7 & 50.4 & 51.6\\
\hline

\end{tabular}}
\label{ntu60results-other}
\end{center}
\vspace{-0.3cm}
\end{table}


Table~\ref{ntu60results-other} shows more evaluations on NTU-60. Before GCNs have become mainstream backbones for the  3D Skeleton-based Action Recognition, encoding 3D body joints of skeletons as texture-like images enjoyed some limited popularity, with approaches \citelatex{supp_ke_2017_CVPR,supp_ke_tip,supp_tas2018cnnbased} feeding such images into CNN backbones. This facilitates easy FSL with existing pipelines such as Matching Nets~\cite{f4Matching} and Prototypical Net~\cite{f1}. Thus, we reshape the normalized 3D coordinates of each skeleton sequence or per block skeleton into image tensors, and pass them into Matching Nets~\cite{f4Matching} and Prototypical Net~\cite{f1} for few-shot learning. Not surprisingly, using texture-like images for skeletons is suboptimal. Our JEANIE is more than 25\% better.


\begin{figure}[tbp!]
\centering
\begin{subfigure}[b]{0.495\linewidth}
\includegraphics[trim=0 0 0 0, clip=true,width=0.99\linewidth]{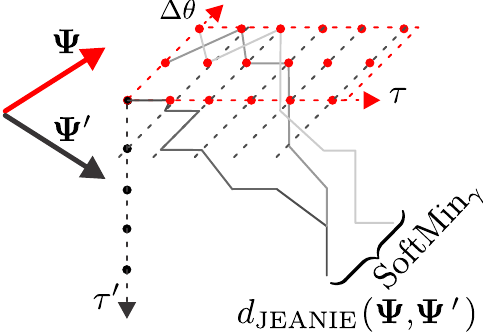}\vspace{-0.1cm}
\caption{\label{fig:jeanie}}
\vspace{-0.3cm}
\end{subfigure}
\begin{subfigure}[b]{0.495\linewidth}
\includegraphics[trim=0 0 0 0, clip=true,width=0.99\linewidth]{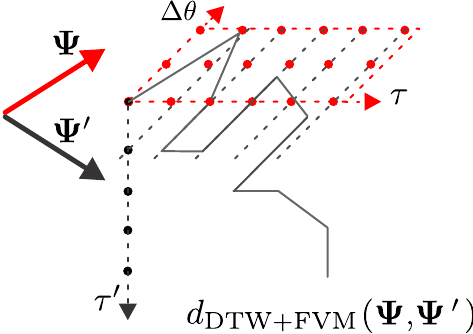}\vspace{-0.1cm}
\caption{\label{fig:fvm}}
\vspace{-0.3cm}
\end{subfigure}
\caption{JEANIE \vs FVM. Fig. \ref{fig:jeanie} shows that in JEANIE, all smooth paths starting from all possible positions  $(\theta_b,1,1)$ and ending at any possible position $(\theta_e,\tau,\tau')$, where $\theta_b,\theta_e\!\in\!\{-\eta\Delta\theta,\cdots,\eta\Delta\theta\}$,  compete to be selected by the SoftMin function. Note these paths change smoothly. In contrast,  Fig. \ref{fig:fvm} shows that in FVM, the path may jump abruptly \wrt $\theta$ as long as the given $\theta$ minimizes locally $d_{fvm}$ used by soft-DTW.
}
\label{fig:jeanie_fvm}
\end{figure}

\section{Free Viewpoint Matching (FVM)} 
\label{fvm}
To ascertain whether our idea, that is, joint temporal and simulated viewpoint alignment (JEANIE) is better than performing separately the temporal and simulated viewpoint alignments, we introduce an important and plausible baseline called the Free Viewpoint Matching (FVM). FVM, for every step of DTW, seeks the best local viewpoint alignment, thus realizing non-smooth temporal-viewpoint path in contrast to JEANIE. To this end, we apply DTW in Eq.~\ref{eq:d_jeanie} with the base distance replaced by:
\begin{align}
& d_{\text{fvm}(\vpsi_{t},\vpsi'_{t'})}\!=\!\softmingg\limits_{m,n\in\{\-\eta,\cdots,\eta\}} d_{\text{base}(\vpsi_{m,n,t},\vpsi'_{m',n',t'})},
\label{eq:suppl1}
\end{align}
where $\mPsi\!\in\!\mbr{d'\times K\times K'\times\tau}$  and $\mPsi'\!\in\!\mbr{d'\times K\times K'\times\tau'}\!$ are query and support feature maps. We abuse slightly the notation by writing $d_{\text{fvm}(\vpsi_{t},\vpsi'_{t'})}$ because we minimize over viewpoint indexes inside of Eq. \eqref{eq:suppl1}. Thus, we evaluate the distance matrix as $\mD\!\in\!\mbrp{\tau\times\tau'}\!\!\equiv\![d_{\text{fvm}}(\vpsi_t,\vpsi'_{t'})]_{(t,t')\in\idx{\tau}\times\idx{\tau'}}$.

Note that this is the FVM variant where we generate multiple candidate viewpoints for both query and support sequences.  Thus, for fair comparison,  JEANIE is also set to go over both azimuth and altitude for both query and support sequences, respectively. Figure \ref{fig:jeanie_fvm} explains key differences between JEANIE and FVM. For brevity, we visualise only one direction of viewpoint alignment \eg, the  azimuth, and we assume in the illustration that we simulate only the multiple query views.

\section{JEANIE \vs FVM }
\label{jeanie_fvm_eval}

Below, we provide our experimental comparisons of JEANIE and FVM.

\begin{table}[tbp!]
\caption{JEANIE \vs FVM on NTU-120.}
\vspace{-0.6cm}
\begin{center}
\resizebox{\linewidth}{!}{\begin{tabular}{ l c | c | c | c | c | c }
\hline
\# Training Classes & & 20  & 40  & 60  & 80  & 100 \\ 
\hline
\hline
APSR& \cite{Liu_2019_NTURGBD120} & 29.1 & 34.8& 39.2 & 42.8 & 45.3 \\
SL-DML& \cite{2021dml}& 36.7 & 42.4 & 49.0 & 46.4& 50.9 \\
Skeleton-DML& \cite{memmesheimer2021skeletondml} & 28.6 & 37.5 & 48.6 & 48.0 & 54.2 \\
\hline
S$^2$GC (baseline, no soft-DTW) & & 30.0 & 35.9 & 39.2 & 43.6 & 46.4  \\ 
soft-DTW (baseline) & & 30.3 & 37.2 & 39.7 & 44.0 & 46.8 \\
{\it FVM}+2V(CamVPC) & & 34.5 & 41.9 & 44.2 & 48.7 & 52.0 \\
{\it JEANIE}+2V(CamVPC) & & 36.8 &  42.6 &  48.3 & 49.8 &  54.9\\
{\it JEANIE}+2V(CamVPC+crossval.) & & {\bf 37.2} & {\bf 43.0} & {\bf 49.2} & {\bf 50.0} & {\bf 55.2}\\ 

\hline
\end{tabular}}
\label{ntu120_fvm_results}
\end{center}
\end{table}

\begin{table*}[tbp!]
\begin{center}
\vspace{-0.3cm}
\caption{Experimental results on the UWA3D Multiview Activity II.
}
\vspace{-0.3cm}
\resizebox{0.75\linewidth}{!}{\begin{tabular}{ l|  c  c  c  c  c c  c  c  c  c c  c  |c  }
\hline
Training view & \multicolumn{2}{c}{$V_1$ \& $V_2$} & \multicolumn{2}{c}{$V_1$ \& $V_3$} & \multicolumn{2}{c}{$V_1$ \& $V_4$} & \multicolumn{2}{c}{$V_2$ \& $V_3$} & \multicolumn{2}{c}{$V_2$ \& $V_4$} & \multicolumn{2}{c|}{$V_3$ \& $V_4$} & Mean\\
\cline{1-13}
Testing view & $V_3$ & $V_4$ & $V_2$ & $V_4$ & $V_2$ & $V_3$ & $V_1$ & $V_4$ & $V_1$ & $V_3$ & $V_1$ & $V_2$ & {}\\
\hline
\hline
S$^2$GC (baseline)&45.5&64.4&46.8&61.6&49.5&43.2&57.3&61.2&51.0&42.9&57.0&49.2&52.5\\
soft-DTW (baseline) &48.2&67.2&51.2&67.0&53.2&46.8&62.4&66.2&57.8&45.0&62.2&53.0&56.7\\
{\it FVM}+2V &50.7&68.8&56.3&69.2&55.8&47.1&63.7&68.8&62.5&51.4&63.8&55.7&59.5\\
{\it JEANIE}+2V&{\bf 55.3}&{\bf 70.2}&{\bf 61.4}&{\bf 72.5}&{\bf 60.9}&{\bf 50.8}&{\bf 66.4}&{\bf 73.9}&{\bf 68.8}&{\bf 57.2}&{\bf 66.7}&{\bf 60.2}&{\bf 63.7}\\
\hline
\end{tabular}}
\label{uwa3dm_fvm_jeanie}
\end{center}
\vspace{-0.2cm}
\end{table*}

\begin{table}[tbp!]
\vspace{-0.3cm}
\caption{JEANIE \vs FVM on NTU-120 (multiview classification). Baseline is soft-DTW and the backbone is S$^2$GC. 
}
\vspace{-0.6cm}
\begin{center}
\resizebox{\linewidth}{!}{\begin{tabular}{ l | c | c | c || c | c  | c }
\hline
Training view & bott. & bott. & bott.\& cent. & left & left & left \& cent.\\
\hline
Testing view & cent. & top & top & cent. & right & right \\
\hline
100/same 100 (baseline)& 74.2 & 73.8 & 75.0 & 58.3 & 57.2 & 68.9\\
100/same 100 ({\it FVM})& 79.9 & 78.2 & 80.0 & 65.9 & 63.9 & 75.0\\

100/same 100 ({\it JEANIE})& {\bf 81.5} & {\bf 79.2} & {\bf 83.9} & {\bf 67.7} & {\bf 66.9} & {\bf 79.2}\\
\hline
100/novel 20 (baseline) & 58.2 & 58.2 & 61.3 & 51.3 & 47.2 & 53.7 \\
100/novel 20 ({\it FVM}) & 66.0 & 65.3 & 68.2 & 58.8 & 53.9 & 60.1 \\
100/novel 20 ({\it JEANIE}) & {\bf 67.8} & {\bf 65.8} & {\bf 70.8} & {\bf 59.5} & {\bf 55.0} & {\bf 62.7} \\
\hline
\end{tabular}}
\label{ntu120results_fvm_view}
\end{center}
\vspace{-0.2cm}
\end{table}

\subsection{One-shot AR on NTU-120}

Table~\ref{ntu120_fvm_results} shows the results on NTU-120 dataset. According to the table, JEANIE outperforms FVM by 2-4\%. This shows that seeking jointly the best temporal-viewpoint alignment is more valuable than considering viewpoint alignment as a local task (free range alignment per each step of soft-DTW). In the table,  the temporal alignment is denoted as ({\em T}), two viewpoint alignments rather than one (\eg, azimuth and altitude) ({\em 2V}), whereas CamVPC means we use the stereo camera geometry to imagine how 3D points would look in a different camera view (rather than Euler angles). Finally, we crossvalidated the azimuth and altitude number of viewpoint steps for ({\em crossval.}).

\subsection{Few-shot AR on  UWA3D Multiview ActivityII.}

Table~\ref{uwa3dm_fvm_jeanie} shows the results on UWA3D Multiview Activity II. Again, JEANIE outperforms  FVM by 4.2\%, and outperforms the baseline (with temporal alignment only) by 7\% on average.

\subsection{Few-shot AR on  NTU-120 (multiview classification).}

Table~\ref{ntu120results_fvm_view} shows our experimental results for the multiview classification protocol (see the main paper for details of this protocol) on NTU-120 dataset. As shown in the table, we see consistent improvement on both eperimental  settings, denoted ({\em 100/same 100})  and ({\em 100/novel 20}). The biggest improvements are observed when there are many different viewpoints being put into the training, \eg, when bottom and center views are used for training and top view for testing, and when left and center views are used for training and the right view is used for testing. This proves the effectiveness of our proposed method in temporal and viewpoint alignment AR.

\comment{\subsection{JEANIE on the Kinetics-skeleton dataset}

Kinetics dataset is a collection of large-scale, high-quality datasets of URL links of up to 650,000 video clips that cover 400/600/700 human action classes, depending on the version of dataset. It includes human-object interactions such as {\it playing instruments}, as well as human-human interactions such as {\it shaking hands} and {\it hugging}. 

As the Kinetics-400 dataset provides only the raw videos, we follow approach~\cite{stgcn2018aaai} and use the estimated joint locations in the pixel coordinate system as the input to our pipeline. To obtain the joint locations, we first resize all videos to the resolution of 340 $\times$ 256, and convert the frame rate to 30 FPS. Then we use the publicly available {\it OpenPose}~\cite{Cao_2017_CVPR} toolbox to estimate the location of 18 joints on every frame of the clips. As OpenPose  provides the 2D body joint coordinates, to obtain 3D coordinates,  multiple cameras or a depth sensor are required. 
As Kinetics-400 does not offer multiview or depth data, we use a network of Martinez \etal   \cite{martinez_2d23d} pre-trained on 
Human3.6M~\cite{Catalin2014Human3}, combined with the 2D OpenPose output to  estimate 3D coordinates from 2D coordinates. While the 2D OpenPose output gives us $(x,y)$ coordinates, the latter network produces $z$ coordinates.

We evaluate our proposed model on both 2D and 3D Kinetics-skeleton datasets. We split the whole dataset into 200 actions for training, and the rest half for testing. As we are unable to estimate the camera location, we simply use Euler angles for the camera viewpoint simulation. 

Table~\ref{kinetics_results} shows the experimental results on this dataset. Using 3D skeletons outperforms the use of 2D skeletons by 3-4\%, and  JEANIE outperforms baseline (temporal alignment only) and FVM by around 5\% and 6\%,  respectively.

\begin{table}
	\centering
	\caption{Experimental results on 2D and 3D Kinetics-skeleton datasets. Note that we have no results on JEANIE or FVM  for 2D coordinates as these require completely different modeling of simulated camera viewpoints (something we can extend our work to in the future).}
	\label{kinetics_results}  
	\vspace{-0.2cm}
	\resizebox{0.9\linewidth}{!}{\begin{tabular}{c|c|c|c|c}  
		\hline  
		&S$^2$GC&S$^2$GC+T&S$^2$GC+T & S$^2$GC+T+2V \\
		& & (baseline) & +2V({\it FVM}) & ({\it JEANIE}) \\%
		\hline
		\hline
		2D skel. & 32.8 & 34.7 & - & - \\
		3D skel. & 35.9 & 39.6 & 44.1 & 50.3 \\
		\hline
	\end{tabular}}
\end{table}}

\section{Evaluation Protocols}
\label{app:epr}

Below, we detail our new/additional evaluation protocols used in the experiments.
\subsection{Few-shot AR protocols on the small-scale datasets}
\label{small_proto}
As we use many class-wise splits for small datasets, these splits will be simply released in our code. Below, we explain the selection process.

\vspace{0.05cm}
\noindent{\bf FSAR (MSRAction3D)}. As this dataset contains 20 action classes, we randomly choose 10 action classes for training and the rest 10 for testing. We repeat this sampling process 10 times to form in total 10 train/test splits. For each split, we have 5-way and 10-way experimental settings. The overall performance on this dataset is computed by averaging the performance on 10 splits.

\vspace{0.05cm}
\noindent{\bf FSAR (3D Action Pairs)}. This dataset has in total 6 action pairs (12 action classes), each pair of action has very similar motion trajectories, \eg, {\it pick up a box} and {\it put down a box}. We randomly choose 3 action pairs to form a training set (6 action classes) and the half action pairs for the test set, and in total there are ${\binom nk}\!=\!{\binom {6}{3}\!=\!20}$ different combinations of train/test splits. As our train/test splits are based on action pairs, we are able to test  whether the algorithm is able to classify unseen action pairs that share similar motion trajectories. We use 5-way protocol on this dataset to evaluate the performance of FSAR, averaged over all 20 splits.

\vspace{0.05cm}
\noindent{\bf FSAR (UWA3D Activity)}. This dataset has 30 action classes. We randomly choose 15 action classes for training and the rest half action classes for testing. We form in total 10 train/test splits, and we use 5-way and 10-way protocols on this dataset, averaged over all 10  splits.

\subsection{One-shot protocol on NTU-60}

Following NTU-120~\cite{Liu_2019_NTURGBD120}, we introduce the one-shot AR setting on NTU-60. We split the whole dataset into two parts: auxiliary set (on NTU-120 the training set is called as auxiliary set, so we follow such a terminology) and one-shot evaluation set. 

\vspace{0.05cm}
\noindent{\bf Auxiliary set} contains 50 classes, and all samples of these classes can be used for learning and validation. Evaluation set consists of 10 novel classes, and one sample from each novel class is picked as the exemplar (terminology introduced by authors of NTU-120), while all the remaining samples of these classes are used to test the recognition performance. 

\vspace{0.05cm}
\noindent{\bf Evaluation set} contains 10 novel classes, namely, A1, A7, A13, A19, A25, A31, A37, A43, A49, A55. The following 10 samples are the exemplars:

\noindent(01)S001C003P008R001A001, (02)S001C003P008R001A007, (03)S001C003P008R001A013, (04)S001C003P008R001A019, (05)S001C003P008R001A025, (06)S001C003P008R001A031, (07)S001C003P008R001A037, (08)S001C003P008R001A043, (09)S001C003P008R001A049, (10)S001C003P008R001A055. 


\vspace{0.05cm}
\noindent{\bf Auxiliary set} contains  50 classes (the remaining 50 classes of NTU-60 excluding the 10 classes in evaluation set).

\subsection{Few-shot multiview classification on NTU-120}

\vspace{0.05cm}
\noindent{\bf Horizontal camera view}.
As NTU-120 is captured by 3 cameras (from 3 different horizontal angles: -45$^{\circ}$, 0$^{\circ}$, 45$^{\circ}$), we split the whole dataset based on the camera ID to form our 3 horizontal camera viewpoints (left, center and right views). We then evaluate few-shot multiview classification using (i) the left view for training and the center view for testing (ii) the left view for training and the right view for testing (ii) the left and center views for training and the right view for testing.

\vspace{0.05cm}
\noindent{\bf Vertical camera view}. Based on the table provided in~\cite{Liu_2019_NTURGBD120}, we first group 32 camera setups into 3 groups by dividing the range of heights into 3 equally-sized ranges to form roughly the top, center and bottom views. We then group the whole dataset into 3 camera viewpoints based on the camera setup IDs. For few-shot multiview classification, we evaluate our proposed method using (i) bottom view for training and center view for testing (ii) bottom view for training and top view for testing (iii) bottom and center views for training and top view for testing.

\section{Network configuration and training details}
\label{network_train}

For reproducibility purposes, \textbf{\color{red} we will release the full code} which also contains evaluation subroutines implementing training and evaluation protocols. Below we provide the details of network configuration and training process in the following sections.

\subsection{Network configuration}

Given the temporal block size $M$ (the number of frames in a block) and desired output size $d$, the configuration of the 3-layer MLP unit is: FC ($3M \rightarrow 6M$), LayerNorm (LN) as in \cite{dosovitskiy2020image}, ReLU, FC ($6M \rightarrow 9M$), LN, ReLU, Dropout (for smaller datasets, the dropout rate is 0.5; for large-scale datasets, the dropout rate is 0.1), FC ($9M \rightarrow d$), LN. Note that $M$ is the temporal block size
~and $d$ is the output feature dimension per body joint. Note that ablations on the value of $M$ are already conducted in Table \ref{blockframe_overlap}.

\noindent\textbf{Backbone with GNN and Transformer.} 
Following EN described in Section \ref{sec:appr}, let us take the query input $\mX\!\in\!\mbr{3\times J\times M}$ for the temporal block of length $M$ as an example, where $3$ indicates 3D Cartesian coordinate and $J$ is the number of body joints. As alluded to earlier, we obtain $\widehat{\mX}^T\!=\!\text{MLP}(\mX; \mathcal{F}_{MLP})\!\in\!\mbr{{d}\times J}$.

Subsequently, we employ a GNN and the transformer encoder \cite{dosovitskiy2020image} which consists of alternating layers of Multi-Head Self-Attention (MHSA) and a feed-forward MLP (two FC layers with a GELU non-linearity between them). LayerNorm (LN) is applied before every block, and residual connections after every block. Each block feature matrix $\widehat{\mX} \in \mathbb{R}^{J \times {d}}$ encoded by GNN (without learnable $\bf \Theta$) is then passed to the transformer. Similarly to the standard transformer, we prepend a learnable vector  
${\bf y}_\text{token}\!\in\!\mbr{1\times {d}}$
to the sequence of block features $\widehat{\mX}$ obtained from GNN,
and we also add the positional embeddings ${\bf E}_\text{pos} \in \mathbb{R}^{(1+J) \times {d}}$ based on the sine and cosine functions (standard in transformers) so that token ${\bf y}_\text{token}$ and each body joint enjoy their own unique positional encoding. 
We obtain  $\mZ_0\!\in\!\mbr{(1+J)\times {d}}$ which is the input in the following backbone:
%
%
\begin{align}
%
&{\bf Z}_0 = [{\bf y}_\text{token}; \text{GNN}(\widehat{\mX})]+{\bf E}_\text{pos}, \label{eq:proj}\\
&{\bf Z}^\prime_k = \text{MHSA}(\text{LN}({\bf Z}_{k-1})) + {\bf Z}_{k-1}, \;k = 1, \cdots, L_\text{tr}\label{eq:mhsa}\\
& {\bf Z}_k = \text{MLP}(\text{LN}({\bf Z}^\prime_k)) + {\bf Z}^\prime_k, \qquad\quad \,k = 1, \cdots, L_\text{tr}\label{eq:mlp} \\
%
& \vy' = \text{LN}\big(\mZ^{(0)}_{L_\text{tr}}\big) \qquad\qquad\text{ where }\quad\;\; \vy'\in \mathbb{R}^{1 \times {d}} \label{eq:blockfeat}\\
&f(\mX; \mathcal{F})=\text{FC}(\vy'^T; \mathcal{F}_{FC})\qquad\qquad\quad\!\in \mathbb{R}^{d'},\label{eq:final_fc_bl}
%
%
\end{align}
where $\mZ^{(0)}_{L_\text{tr}}$ is the first ${d}$ dimensional row vector extracted from the output matrix $\mZ_{L_\text{tr}}$  of size $(J\!+\!1)\times{d}$ which corresponds to the last layer $L_\text{tr}$ of the transformer. Moreover, parameter $L_\text{tr}$ controls the depth of the transformer, whereas $\mathcal{F}\!\equiv\![\mathcal{F}_{MLP},\mathcal{F}_{GNN},\mathcal{F}_{Transf},\mathcal{F}_{FC}]$ is the  set of  parameters of EN. In case of APPNP, SGC and S$^2$GC, $|\mathcal{F}_{GNN}|\!=\!0$ because we do not use their learnable parameters $\bf \Theta$ (\ie, think $\bf \Theta$ is set as the identity matrix).

As in Section \ref{sec:appr}, one can define now a support feature map as $\mPsi'\!=\![f(\boldsymbol{X}_1;\mathcal{F}),\cdots,f(\boldsymbol{X}_{\tau'};\mathcal{F})]\!\in\!\mbr{d'\times\tau'}$ for $\tau'$ temporal blocks, and the query map $\mPsi$ accordingly.

 The hidden size of our transformer  (the output size of the first FC layer of the MLP in Eq. \eqref{eq:mlp}) depends on the dataset. For smaller datasets, the depth of the transformer is $L_\text{tr}\!=\!6$ with $64$ as the hidden size, and the MLP output size is  ${d}\!=\!32$ (note that the MLP which provides $\widehat{\mX}$ and the MLP in the transformer must both have  the same output size). For NTU-60, the depth of the transformer is $L_\text{tr}\!=\!6$, the hidden size is 128 and the MLP output size is  ${d}\!=\!64$. For NTU-120, the depth of the transformer is   $L_\text{tr}\!=\!6$, the hidden size is 256 and the MLP size is ${d}\!=\!128$. For Kinetics-skeleton, the depth for the transformer is $L_\text{tr}\!=\!12$, hidden size is 512 and the MLP output size is ${d}\!=\!256$. The number of Heads for the transformer of smaller datasets, NTU-60, NTU-120 and Kinetics-skeleton is set as 6, 12, 12 and 12, respectively.

The output sizes $d'$ of the final FC layer in Eq. \eqref{eq:final_fc_bl} are 50, 100, 200, and 500 for the smaller datasets, NTU-60, NTU-120 and Kinetics-skeleton, respectively.


\subsection{Training details}
The weights for the pipeline are initialized with the normal distr. (zero mean and unit standard dev.).
We use 1e-3 for the learning rate, and the weight decay is 1e-6. We use the SGD optimizer.
We set the number of training episodes to 100K for NTU-60, 200K for NTU-120, 500K for 3D Kinetics-skeleton, 10K for small datasets such as UWA3D Multiview Activity II.                        
We use Hyperopt for hyperparam. search on validation sets for all the datasets.

\section{Skeleton Data Preprocessing}

Before passing the skeleton sequences into MLP and graph networks (\eg, S$^2$GC), we first normalize each body joint \wrt to the torso joint ${\bf v}_{f, c}$:
\begin{equation}
    {\bf v}^\prime_{f, i}\!=\!{\bf v}_{f, i}\!-\!{\bf v}_{f, c},
\end{equation}
where $f$ and $i$ are the index of video frame and human body joint respectively. After that, we further normalize each joint coordinate into  [-1, 1] range:

\begin{equation}
    \hat{{\bf v}}_{f, i}[j] = \frac{{\bf v}^\prime_{f, i}[j]}{ \text{max}([\text{abs}({\bf v}^\prime_{f, i}[j])]_{f\in\idx{\tau},i\in\idx{J} } )},
\end{equation}
where $j$  selects the $x$, $y$ and $z$ axes, $\tau$ is the number of frames and $J$ is the number of 3D body joints per frame.

For the skeleton sequences with more than one performing subject, (i) we normalize each skeleton separately, and each skeleton is passed to MLP for learning the temporal dynamics, and (ii) for the output features per skeleton from MLP, we pass them separately to graph networks, \eg, two skeletons from a given video sequence will have two outputs from the graph networks, and we aggregate the outputs by average pooling before passing to FVM or JEANIE.


\bibliographystylelatex{ieee_fullname}
\bibliographylatex{jeanie}

\end{document}